\documentclass[10pt,twocolumn,letterpaper]{article}

\usepackage{cvpr}
\usepackage{times}
\usepackage{epsfig}
\usepackage{graphicx}
\usepackage{amsmath}
\usepackage{amssymb}

\def\R{{\rm I} \! {\rm R}}

\newcommand{\SKIP}[1]{} 

\newcommand{\mbegin} {\left [ \begin{array}}

\newcommand{\mend}   {\end{array} \right ]}

\newcommand{\detbegin} {\left | \begin{array}}
\newcommand{\detend}   {\end{array} \right |}

\newcommand{\vbegin} {\left ( \begin{array}{c}}

\newcommand{\vend} {\end{array}\right )}

\def\squareforqed{\hbox{\rlap{$\sqcap$}$\sqcup$}}

\def\qed{\ifmmode\squareforqed\else{\unskip\nobreak\hfil
	\penalty50\hskip1em\null\nobreak\hfil\squareforqed
	\parfillskip=0pt\finalhyphendemerits=0\endgraf}\fi}

\newcommand{\showeqnlabel}{
	\hbox to 0pt{\quad\quad\relax\fbox{\scriptsize\rm\eqnlblx}%
	\gdef\eqnlblx{xxxx}}} \newcommand{\eqnlblx}{}

\def\@eqnnum{\rm (\theequation)\showeqnlabel}

\newcommand{\nofig}[1]{\centerline{\bf Figure here}}

\usepackage[pagebackref=true,breaklinks=true,letterpaper=true,colorlinks,bookmarks=false]{hyperref}
\usepackage{multicol}
\usepackage{caption}
\usepackage{multirow}
\newcommand{\comment}[1]{}

\cvprfinalcopy 

\def\cvprPaperID{4801} 

\ifcvprfinal\fi

\begin{document}

\twocolumn[{
\begin{@twocolumnfalse}
\title{Cost Volume Pyramid Based Depth Inference for Multi-View Stereo}

\author{Jiayu Yang$^1$, \;\;Wei Mao$^1$,\;\; Jose M. Alvarez$^2$,\;\; Miaomiao Liu$^{1,3}$\\
$^1$Australian National University, $^2$NVIDIA, $^3$Australian Centre for Robotic Vision\\
{\tt\small \{jiayu.yang, wei.mao, miaomiao.liu\}@anu.edu.au,}\;\;{\tt\small josea@nvidia.com}
}

\maketitle
\thispagestyle{empty}

\begin{center}
\setlength\tabcolsep{1pt}
\vspace{-0.55cm}\begin{tabular}{cccc}
      \includegraphics[width=0.32\linewidth]{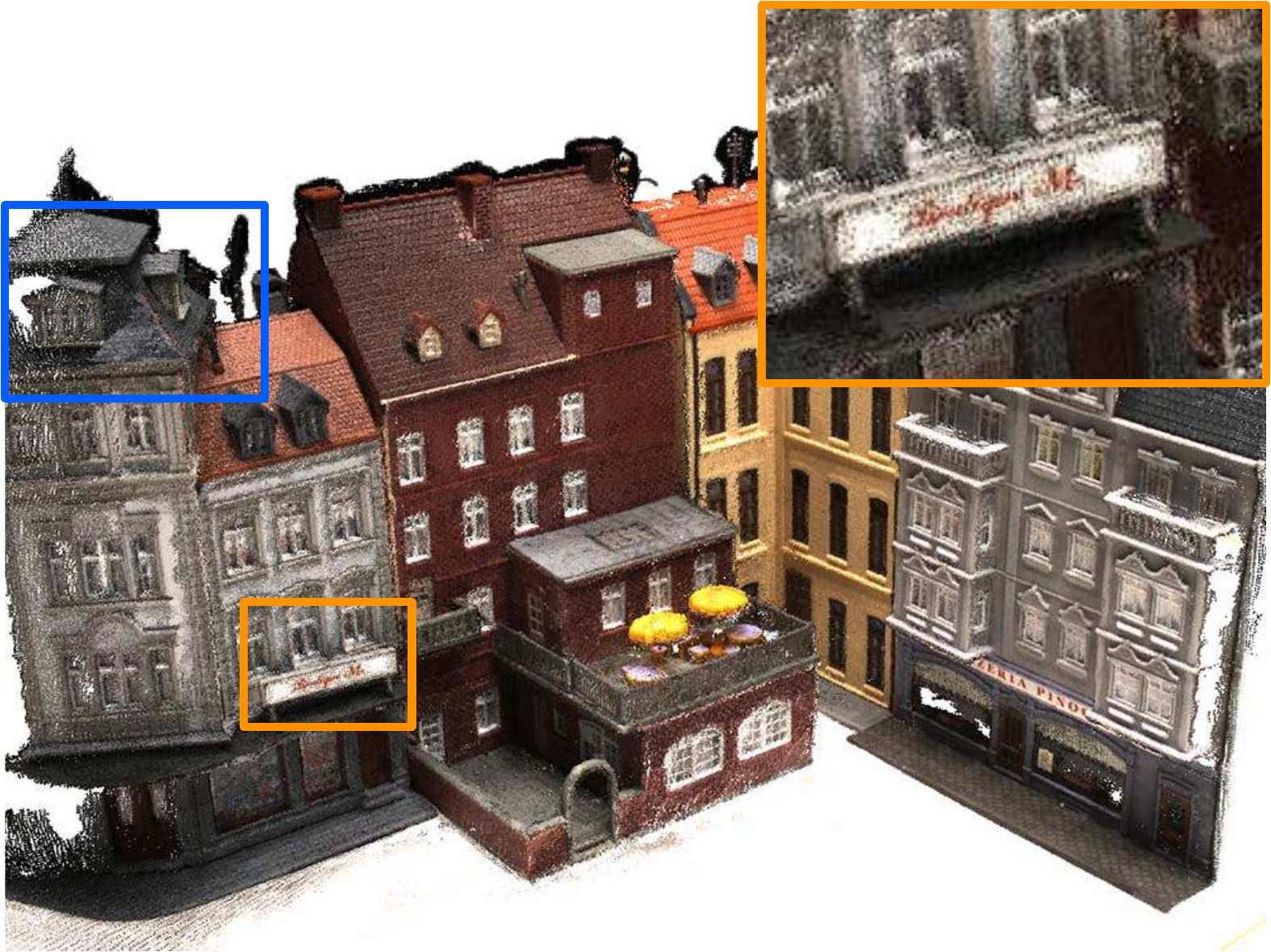}
      & \includegraphics[width=0.32\linewidth]{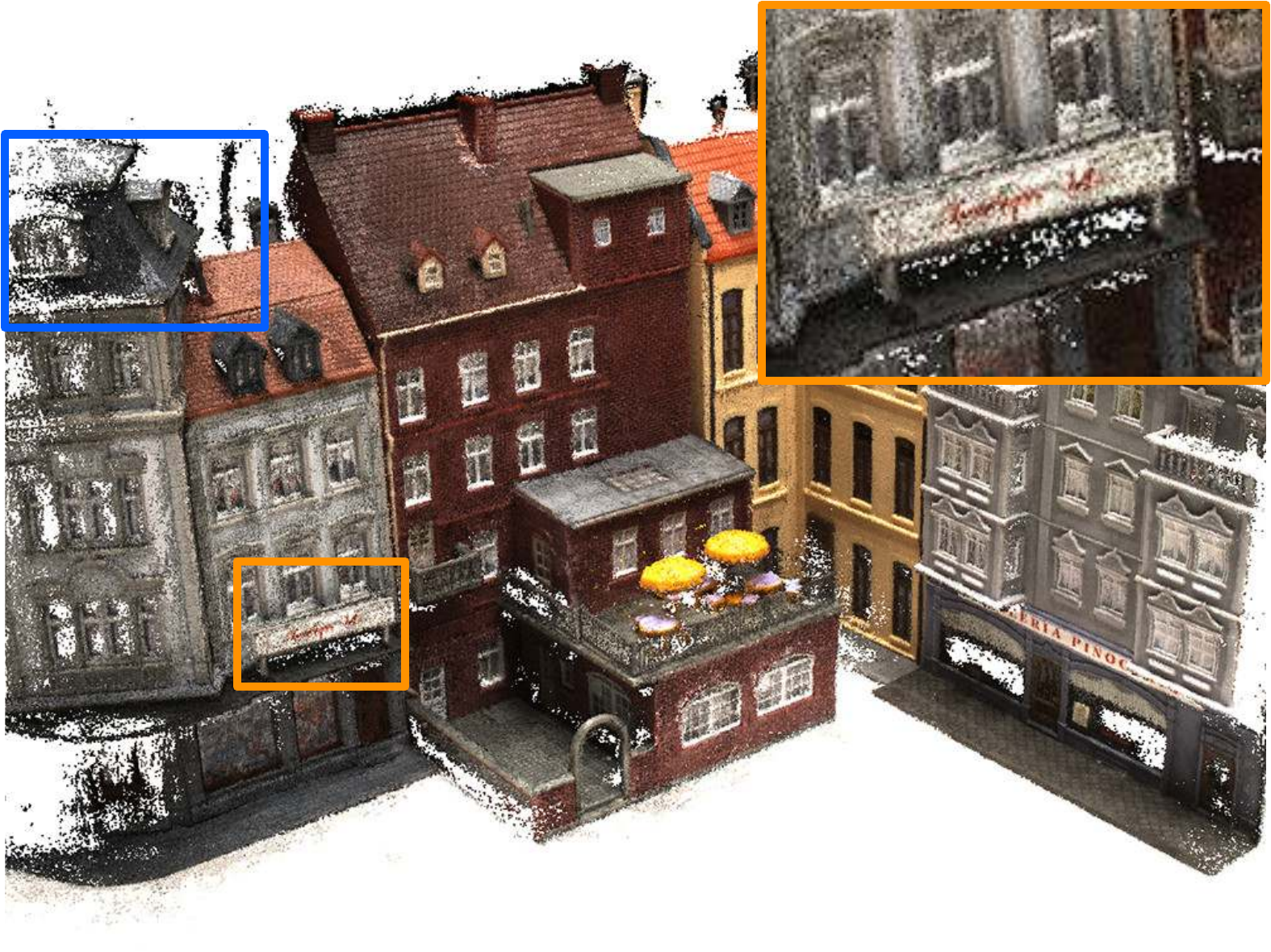}
      & \includegraphics[width=0.32\linewidth]{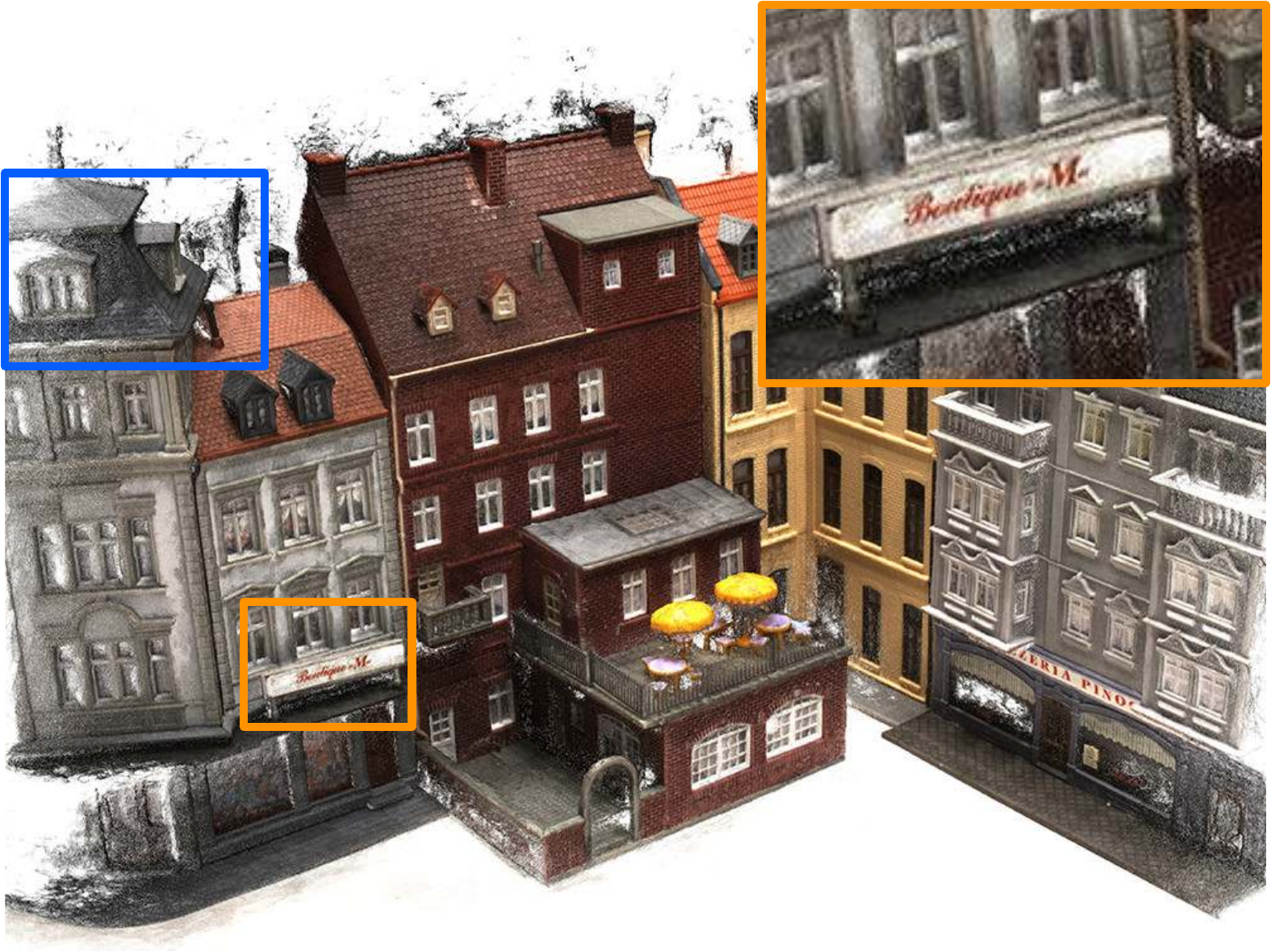}\\
      R-MVSNet~\cite{yao2019recurrent} & Point-MVSNet~\cite{chen2019point} & Ours
\end{tabular}
\end{center}
\vspace{-0.5cm}
\captionof{figure}{Point clouds reconstructed by state-of-the-art methods~\cite{chen2019point,yao2019recurrent} and our CVP-MVSNet. Best viewed on screen.\vspace{0.3cm} }\label{fig:startfig}
\end{@twocolumnfalse}
}]

\vspace{+0.2cm}
\begin{abstract}

\vspace{-0.2cm}
We propose a cost volume-based neural network for depth inference from multi-view images.~We demonstrate that building a~\emph{cost volume pyramid} in a coarse-to-fine manner instead of constructing a cost volume at a fixed resolution leads to a compact, lightweight network and allows us inferring high resolution depth maps to achieve better reconstruction results.~To this end, we first build a cost volume based on uniform sampling of fronto-parallel planes across the entire depth range at the coarsest resolution of an image. Then, given current depth estimate, we construct new cost volumes iteratively on the pixelwise depth residual to perform depth map refinement. While sharing similar insight with~\emph{Point-MVSNet} as predicting and refining depth iteratively, we show that working on~\emph{cost volume pyramid} can lead to a more compact, yet efficient network structure compared with the~\emph{Point-MVSNet} on 3D points.~We further provide detailed analyses of the relation between (residual) depth sampling and image resolution, which serves as a principle for building compact~\emph{cost volume pyramid}.~Experimental results on benchmark datasets show that our model can perform 6\emph{x} faster and~has similar performance as state-of-the-art methods. Code is available at \url{https://github.com/JiayuYANG/CVP-MVSNet}.

\end{abstract}
\vspace{-0.5cm}
\section{Introduction}
Multi-view stereo (MVS) aims to reconstruct the 3D model of a scene from a set of images captured by a camera from multiple viewpoints. It is a fundamental problem for computer vision community and has been studied extensively for decades~\cite{seitz2006comparison}.~While traditional methods before deep learning era have great achievements on the reconstruction of a scene with Lambertian surfaces, they still suffer from illumination changes, low-texture regions, and reflections resulting in unreliable matching correspondences for further reconstruction.

Recent learning-based approaches~\cite{chen2019point, yao2018mvsnet,yao2019recurrent} adopt deep CNNs to infer the depth map for each view followed by a separate multiple-view fusion process for building 3D models.~These methods allow the network to extract discriminative features encoding global and local information of a scene to obtain robust feature matching for MVS. In particular, Yao \etal propose MVSNet~\cite{yao2018mvsnet} to infer a depth map for each view. An essential step in~\cite{yao2018mvsnet} is to build a cost volume based on a plane sweep process followed by multi-scale 3D CNNs for regularization.~While effective in depth inference accuracy, its memory requirement is cubic to the image resolution. To allow handling high resolution images, they then adopt a recurrent cost volume regularization process~\cite{yao2019recurrent}. However, the reduction in memory requirements involves a longer run-time.

In order to achieve a computationally efficient network, Chen \etal~\cite{chen2019point} work on 3D point clouds to iteratively predict the depth residual along visual rays using~\emph{edge~convolutions} operating on the $k$ nearest neighbors of each 3D point. While this approach is efficient, its run-time increases almost linearly with the number of iteration levels.

In this work, we propose a~\emph{cost volume pyramid} based Multi-View Stereo Network (CVP-MVSNet) for depth inference. In our approach, we first build an image pyramid for each input image. Then, for the coarsest resolution of the reference image, we build a compact cost volume by sampling the depth across the entire depth-range of a scene. After that, at the next pyramid level, we perform residual depth search from the neighbor of the current depth estimate to construct a~\emph{partial cost volume} using multi-scale 3D CNNs for regularization. As we build these cost volumes iteratively with a short search range at each level, it leads to a small and compact network. As a result, our network performs 6x faster than current state-of-the-art networks on benchmark datasets.
 
While it is noteworthy that we share the similar insight with~\cite{chen2019point} as predicting and refining the depth map in a coarse-to-fine manner, our work differs from theirs in the following four main aspects. First, the approach in~\cite{chen2019point} performs convolutions on 3D point cloud. Instead, we construct cost volumes on a regular grid defined on the image coordinates, which is shown to be faster in run-time. Second, we provide a principle for building a compact~\emph{cost volume pyramid} based on the correlation between depth sampling and image resolution. As third main difference, we use a multi-scale 3D-CNN regularization to cover large receptive field and encourage local smoothness on residual depth estimates which, as shown in Fig.~\ref{fig:startfig}, leads to a better accuracy. Finally, in contrast to~\cite{chen2019point} and other related works, our approach can output depth of small resolution with small resolution image.  

In summary, our main contributions are
\begin{itemize}
	\item We propose a cost-volume based, compact, and computational efficient depth inference network for MVS.
	\item We build a~\emph{cost volume pyramid} in a coarse-to-fine manner based on a detailed analysis of the relation between the depth residual search range and the image resolution,
	\item Our framework can handle high resolution images with less memory requirement, is 6x faster than the current state-of-the-art framework, i.e.~Point-MVSNet~\cite{chen2019point}, and achieves a better accuracy on benchmark datasets.
\end{itemize}

\begin{figure*}
	\begin{center}
\includegraphics[width=\linewidth]{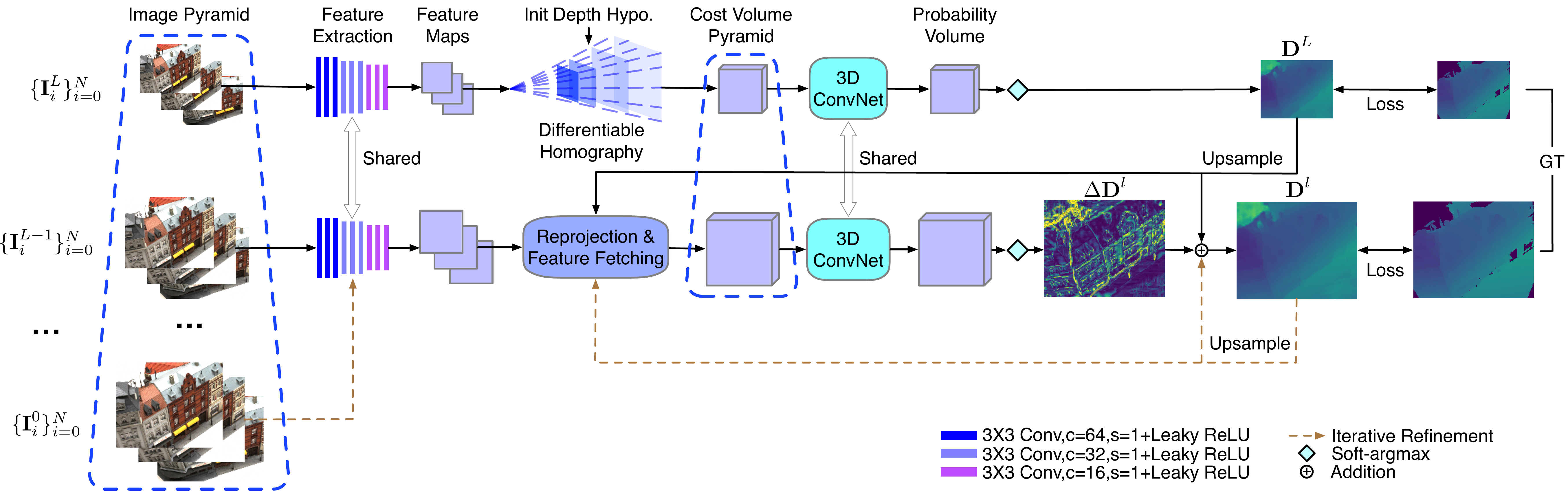}
\vspace{+0.03cm}
	\caption{\label{fig:eth} 
	Network Structure. Reference and source images are first downsampled to form an image pyramid. We apply~\emph{feature extraction network} to all levels and images to extract feature maps. We then build the~\emph{cost volume pyramid} in a coarse-to-fine manner. Specifically, we start with the construction of a cost volume corresponding to coarsest image resolution followed by building~\emph{partial cost volumes} iteratively for depth residual estimation in order to achieve depth map ${\bf D}={\bf D}^0$ for ${\bf I}$. Please refer to Fig.~\ref{fig:reproj} for details about re-projection, feature fetching and building cost volume.
	}
	\label{fig:networkStructure}
	\end{center}
	\vspace{-0.77cm}
\end{figure*}
\section{Related Work}
\noindent{\bf Traditional Multi-View Stereo.}~Multi-view stereo has been extensively studied for decades.~We refer to algorithms before deep learning era as traditional MVS methods which represent the 3D geometry of objects or scene using voxels~\cite{de1999poxels,kutulakos2000theory}, level-sets~\cite{faugeras2002variational,pons2003variational}, polygon meshes~\cite{esteban2004silhouette,fua1995object} or depth maps~\cite{kang2001handling,kolmogorov2002multi}. In the following, we mainly focus on discussions about volumetric and depth-based MVS methods which have been integrated to learning-based framework recently. 

Volumetric representations can model most of the objects or scenes.~Given a fixed volume of an object or scene, volumetric-based methods first divide the whole volume into small voxels and then use a photometric consistency metric to decide whether the voxel belongs to the surface or not. 
These methods do not impose constraints on the shape of the objects. However, the space discretisation is memory intensive. By contrast, depth-map based MVS methods have shown more flexibility in modeling the 3D geometry of scene~\cite{merrell2007real}.~Readers are referred to~\cite{seitz2006comparison} for detailed discussions. Similar to other recent learning-based approaches, we adopt depth map representation in our framework.

\vspace{0.1cm}
\noindent{\bf Deep learning-based MVS.} Deep CNNs have significantly advanced the progress of high-level vision tasks, such as image recognition~\cite{he2016deep,simonyan2014very}, object detection~\cite{girshick2015fast, ren2015faster}, and semantic segmentation~\cite{chen2017deeplab,long2015fully}. As for 3D vision tasks, learning-based approaches have been widely adopted to solve stereo matching problems and have achieved very promising results~\cite{batsos2018cbmv,luo2016efficient,zbontar2016stereo}. However, these learning-based approaches cannot be easily generalized to solve MVS problems as rectifications are required for the multiple view scenario which may cause the loss of information~\cite{yao2018mvsnet}.

More recently, a few approaches have proposed to directly solve MVS problems~\cite{Hartmann2017LearnedMS,leroy2018shaperecon,paschalidou2019raynet}. For instance, Ji~\etal ~\cite{ji2017surfacenet} introduce the first learning based pipeline for MVS. This approach learns the probability of voxels lying on the surface. Concurrently, Kar~\etal~\cite{kar2017learning} present a learnable system to up-project pixel features to the 3D volume and classify whether a voxel is occupied or not by the surface. These systems provide promising results. However, they use volumetric representations that are memory expensive and therefore, these algorithms can not handle large-scale scenes.

Large-scale scene reconstruction has been approached by Yao \etal in~\cite{yao2018mvsnet}. The authors propose to learn the depth map for each view by constructing a cost volume followed by 3D CNN regularization.~Then, they obtain the 3D geometry by fusing the estimated depth maps from multiple views.~The algorithm uses cost volume with memory requirements cubic to the image resolution.~Thus, it can not leverage all the information available in high-resolution images.~To circumvent this problem, the algorithm adopts GRUs~\cite{yao2019recurrent} to regularize the cost volume in a sequential manner. As a result, the algorithm reduces the memory requirement but leads to increased run-time.

Closely related work to ours is Point-MVSNet~\cite{chen2019point}. Point-MVSNet is a framework to predict the depth in a coarse-to-fine manner working directly on point cloud.~It allows the aggregation of information from its $k$ nearest neighbors in 3D space. Our approach shares similar insight as predicting and refining depth maps iteratively.~However, we differ from Point-MVSNet in a few key aspects: Instead of working on 3D, we build the cost volume on the regular image grid. Inspired by the idea of~\emph{partial cost volume} used in PWC-Net~\cite{sun2018pwc} for optical flow estimation, we build ~\emph{partial cost volume} to predict depth residuals.~We compare the memory and computational efficiency with~\cite{chen2019point}. It shows that our cost-volume pyramid based network leads to more compact and accurate models that run much faster for a given depth-map resolution.

\vspace{0.2cm}
\noindent{\bf Cost volume.} Cost volume is widely used in traditional methods for dense depth estimation from unrectified images~\cite{collins1996space, yang2003multi}.
However, most recent learning-based works build cost volume at a fixed resolution~\cite{im2018dpsnet,yao2018mvsnet,yao2019recurrent}, which leads to high memory requirement for handling high resolution images.~Recently, Sun \etal~\cite{sun2018pwc} introduced the idea of~\emph{partial cost volume} for optical flow estimation. In short, given an estimate of the current optical flow, the~\emph{partial cost volume} is constructed by searching correspondences within a rectangle around its position locally in the~\emph{warped} source view image. Inspired by such strategy, in this paper, we propose~\emph{cost volume pyramid} as an algorithm to progressively estimate the depth residual for each pixel along its visual ray. As we will demonstrate in our experiments, constructing cost volumes at multiple levels leads to a more effective and efficient framework.

\vspace{-0.3cm}
\section{Method}
Let us now introduce our approach to depth inference for MVS. The overall system is depicted in Fig.~\ref{fig:networkStructure}. As existing works, we assume the reference image is denoted as ${\bf I}_0\in \mathbb{R}^{H\times W}$, where $H$ and $W$ define its dimensions. Let $\{{\bf I}_i\}_{i=1}^N$ be its $N$ neighboring source images.~Assume $\{{\bf K}_i, {\bf R}_i, {\bf t}_i\}_{i=0}^N$ are the corresponding camera intrinsics, rotation matrix, and translation vector for all views.  Our goal is to infer the depth map ${\bf D}$ for ${\bf I}_0$ from $\{{\bf I}_i\}_{i=0}^N$. The key novelty of our approach is using a feed-forward deep network on~\emph{cost volume pyramid} constructed in a coarse-to-fine manner.~Below, we introduce our feature pyramid, the~\emph{cost volume pyramid}, depth map inference and finally provide details of the loss function.

\subsection{Feature Pyramid}
As raw images vary with illumination changes, we adopt learnable features, which has been demonstrated to be crucial step for extracting dense feature correspondences~\cite{yao2018mvsnet,tang2018banet}. The general practice in existing works is to make use of high resolution images to extract multi-scale image features even for the output of a low resolution depth map. By contrast, we show that a low resolution image contains enough information useful for estimating a low resolution depth map.

Our feature extraction pipeline consists of two steps, see Fig.~\ref{fig:networkStructure}. First, we build the ($L+1$)-level image pyramid $\{{\bf I}^j_i\}_{j=0}^L$ for each input image, $i\in\{0,1,\cdots,N\}$, where the bottom level of the pyramid corresponds to the input image, ${\bf I}^0_i={\bf I}^{}_i$. Second, we obtain feature representations at the $l$-th level using a CNN, namely~\emph{feature extraction network}. Specifically, it consists of 9 convolutional layers, each of which is followed by a leaky rectified linear unit (Leaky-ReLU). We use the same CNN to extract features for all levels in all the images.~We denote the feature maps for a given level $l$ by $\{{\bf f}_i^l\}_{i=0}^N, {\bf f}_i^l\in \mathbb{R}^{H/2^l\times W/2^l \times F}$, where $F=16$ is the number of feature channels used in our experiments. We will show that, compared to existing works, our feature extraction pipeline leads to significant reduction in memory requirements and, at the same time, improve performance.

\vspace{-0.14cm}
\subsection{Cost Volume Pyramid}
Given the extracted features, the next step is to construct cost volumes for depth inference in the reference view. Common approaches usually build a single cost volume at a fixed resolution~\cite{im2018dpsnet,yao2018mvsnet,yao2019recurrent}, which incurs in large memory requirements and thus, limits the use of high-resolution images. Instead, we propose to build a~\emph{cost volume pyramid}, a process that iteratively estimates and refines depth maps to achieve high resolution depth inference. More precisely, we first build a cost volume for coarse depth map estimation based on images of the coarsest resolution in image pyramids and uniform sampling of the fronto-parallel planes in the scene. Then, we construct partial cost volumes based on coarse estimation and depth residual hypotheses iteratively to achieve depth maps with higher resolution and accuracy. We provide details about these two steps below.

\begin{figure}[!t]
	\begin{center}
\includegraphics[width=1.0\linewidth]{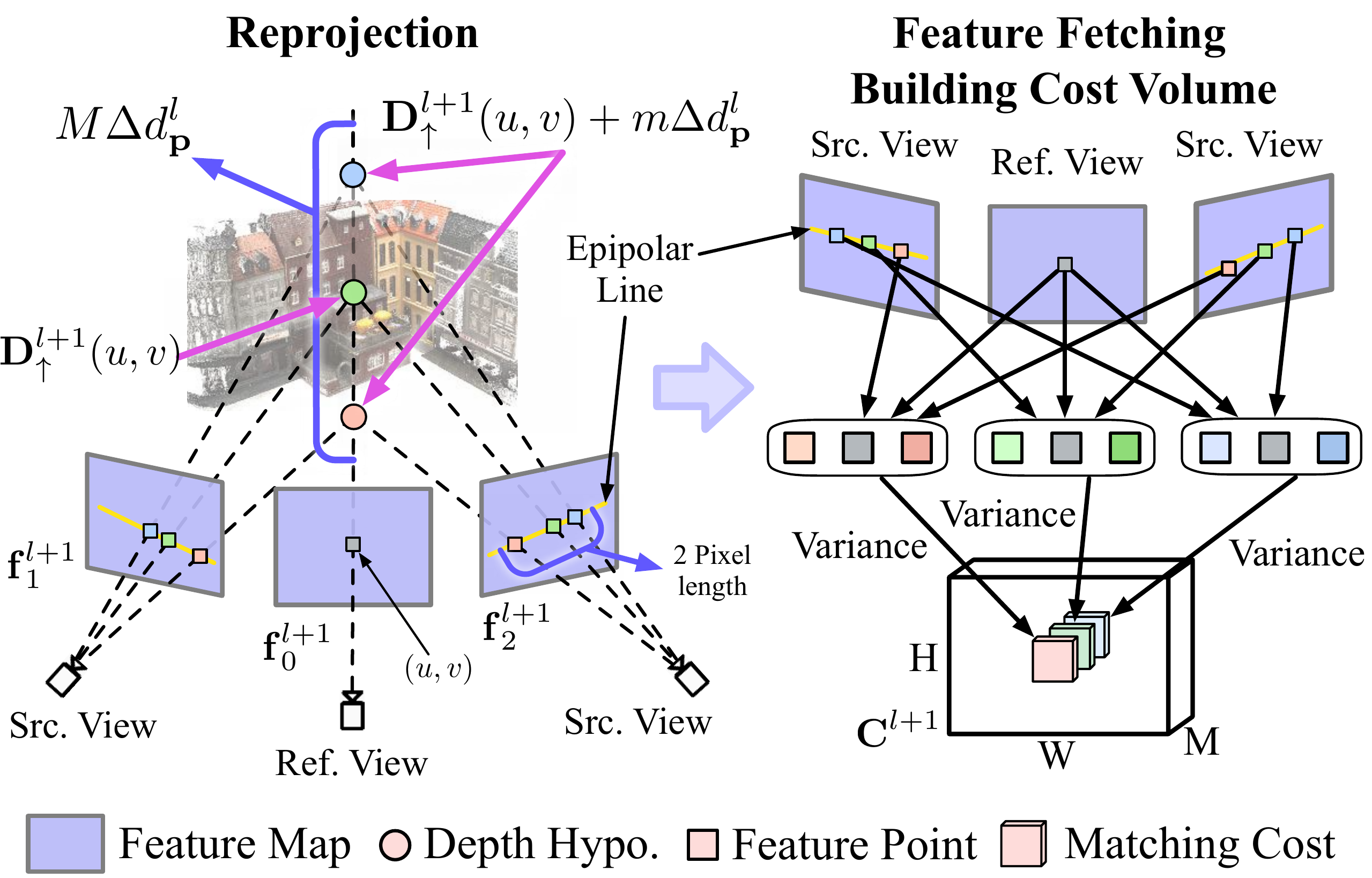}
	\caption{Reprojection, feature fetching and building cost volume. Left: We define $M$ depth hypotheses for each pixel ($u,v$) in the reference view. By projecting them to each source view, we can fetch $M$ corresponding features.~Right: For each depth hypothesis, the matching cost is the variance of fetched features across source views and the reference view. The cost volume ${\bf C}^{l+1}$ defines matching costs for all depth hypotheses of all pixels in the reference view.}
	\label{fig:reproj}
	\end{center}
	\vspace{-0.7cm}
\end{figure}
\noindent{\bf Cost Volume for Coarse Depth Map Inference.}~We start building a cost volume at the $L$th level corresponding to the lowest image resolution $(H/2^L, W/2^L)$.~Assume depth measured at the reference view of a scene ranges from $d_{\text{min}}$ to $d_{\text{max}}$. We construct the cost volume for the reference view by sampling $M$ fronto-parallel planes uniformly across entire depth range. A sampled depth $d =d_{\text{min}}+m (d_{\text{max}} - d_{\text{min}})/M, m\in\{0,1,2,\cdots,M-1\}$ represents a plane where its normal ${\bf n}_0$ is the principal axis of the reference camera.

Similar to~\cite{yao2018mvsnet}, we define the differentiable homography ${\bf H}_i(d)$ between the $i$th source view and the reference view at depth $d$ as
\vspace{-0.3cm}
\begin{equation}
	{\bf H}_i(d) = {\bf K}_i^L{\bf R}_i({\bf I} - \frac{({\bf t}_0-{\bf t}_i){\bf n}_0^T}{d}){\bf R}_0^{-1}({\bf K}_0^L)^{-1},
	\label{eq:dhomo}
\end{equation}
where $\bf I$ is the identity matrix, and ${\bf K}_i^L$ and ${\bf K}_0^L$ are the scaled intrinsic matrices of ${\bf K}_i$ and ${\bf K}_0$ at level $L$. 

Each homography ${\bf H}_i(d) $ suggests a possible pixel correspondence between $\tilde{\bf x}_i$ in source view $i$ and a pixel ${\bf x}$ in the reference view. This correspondence is defined as $\lambda_i\tilde{\bf x}_i={\bf H}_i(d){\bf x}$, where $\lambda_i$ represents the depth of $\tilde{\bf x}_i$ in the source view $i$. 

Given $\tilde{\bf x}_i$ and $\{{\bf f}_i^L\}_{i=1}^N$, we use differentiable bilinear interpolation to reconstruct the feature map warped to the reference view $\{\tilde{\bf f}_{i,d}^L\}_{i=1}^N$. The cost for all pixels at depth $d$ is defined as its variance of features from $N+1$ views,
 \vspace{-0.3cm}
 \begin{equation}
 	{\bf C}_d^L= \frac{1}{(N+1)}\sum_{i=0}^{N}(\tilde{\bf f}_{i,d}^L - \bar{{\bf f}}_d^L )^2,
 	\label{eq:vard}
 	\vspace{-0.3cm}
 \end{equation}
where $\tilde{\bf f}_{0,d}^L={\bf f}_{0}^L$ is the feature map of the reference image and $\bar{{\bf f}}_d^L$ is the average of feature volumes across all views ($\{\tilde{\bf f}_{i,d}^L\}_{i=1}^N\cup {\bf f}_0^L$) for each pixel.~This metric encourages that the correct depth for each pixel has the smallest feature variance, which corresponds to the photometric consistency constraint.~We compute the cost map for each depth hypothesis and concatenate those cost maps to a single cost volume ${\bf C}^L\in \R^{W/2^L\times H/2^L\times M\times F}$.

A key parameter to obtain good depth estimation accuracy is the depth sampling resolution $M$.~We will show in~Section~\ref{sec:depthEstimator} how to determine the interval for depth sampling and coarse depth estimation.

\vspace{0.1cm}
\noindent{\bf Cost Volume for Multi-scale Depth Residual Inference.}
Recall that our ultimate goal is to obtain ${\bf D}={\bf D}^{0}$ for ${\bf I}_{0}$. We iterate starting from ${\bf D}^{l+1}$, a given depth estimate for the $(l+1)$th level, to obtain a refined depth map for the next level ${\bf D}^l$ until reaching the bottom level. More precisely, we first upsample ${\bf D}^{l+1}$ to the next level ${\bf D}^{l+1}_{\uparrow}$ via bicubic interpolation and then, we build the~\emph{partial cost volume} to regress the~\emph{residual depth map} defined as $\Delta {\bf D}^l$ to obtain a refined depth map ${\bf D}^l={\bf D}^{l+1}_{\uparrow}+\Delta {\bf D}^l$ at the $l$th level. 

While we share the similar insight with~\cite{chen2019point} to iteratively predict the depth residual, we argue that instead of performing convolutions on a point cloud~\cite{chen2019point}, building the regular 3D cost volume on the depth residual followed by multi-scale 3D convolution can lead to a more compact, faster, and higher accuracy depth inference. Our motivation is that depth displacements for neighboring pixels are correlated which indicates that regular multi-scale 3D convolution would provide useful contextual information for depth residual estimation. We therefore arrange the depth displacement hypotheses in a regular 3D space and compute the cost volume as follows. 
\begin{figure}[!ht]
    \hspace{0.2cm}
    \vspace{-0.25cm}
    \begin{center}
    \includegraphics[width=0.8\linewidth]{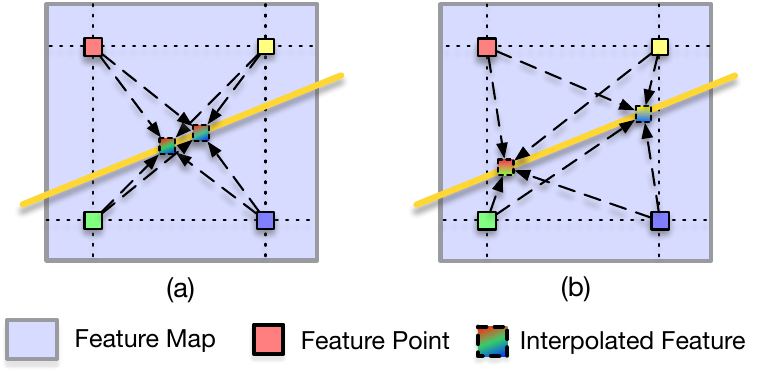}
    \vspace{-0.52cm}
    \end{center}
    \caption{Interpolation of two sampling points from four feature points in source view. (a) Densely sampled depth will result in very close ($<0.5$ pixel) locations which have similar feature. (b) Points projected using appropriate depth sampling carry distinguishable information.}
    \vspace{-0.35cm}
    \label{fig:planeSampling}
\end{figure}

Assume we are given camera parameters $\{{\bf K}^l_i, {\bf R}_i, {\bf t}_i\}_{i=0}^N$ for all camera views and the upsampled depth estimate ${\bf D}^{l+1}_{\uparrow}$.
Current depth estimate for each pixel ${\bf p} = (u,v)$ is defined as $d_{\bf p}={\bf D}^{l+1}_{\uparrow}(u,v)$. Let each depth residual hypothesis interval be $\Delta d_{\bf p} = s_{\bf p}/M$, where $ s_{\bf p}$ represents the depth search range at ${\bf p}$ and $M$ denotes the number of sampled depth residual. We consider the projection of corresponding hypothesized 3D point with depth (${\bf D}^{l+1}_{\uparrow}(u,v)+m\Delta d_{\bf p}$) into view $i$ as %
\begin{equation}
	\resizebox{0.9\linewidth}{!}{$\lambda_i{\bf x}'_i = {\bf K}^l_i({\bf R}_i{\bf R}_0^{-1}(({\bf K}^l_0)^{-1}(u,v,1)^T(d_{\bf p}+m\Delta d_{\bf p})-{\bf t}_0)+{\bf t}_i)$},
\end{equation}
where $\lambda_i$ denotes the depth of corresponding pixel in view $i$, and $m\in\{-M/2,\cdots,M/2-1\}$ (see Fig.~\ref{fig:reproj}). Then, the cost for that pixel at each depth residual hypothesis is similarly defined based on Eq.~\ref{eq:vard}, which leads to a partial cost volume ${\bf C}^l\in\R^{H/2^l\times W/2^l\times M\times F}$. 

In the next section, we introduce our solution to determine the depth search intervals and range for all pixels, $s_{\bf p}$, which is essential to obtain accurate depth estimates.
\subsection{Depth Map Inference}\label{sec:depthEstimator}
In this section, we first provide details to perform depth sampling at the coarsest image resolution and discretisation of the local depth search range at higher image resolution for building the cost volume. Then, we introduce depth map estimators on cost volumes to achieve the depth map inference.

\vspace{0.1cm}
\noindent{\bf Depth Sampling for Cost Volume Pyramid}
We observe that the depth sampling for virtual depth planes is related to the image resolution. As shown in Fig.~\ref{fig:planeSampling}, it is not necessary to sample depth planes densely as projections of those sampled 3D points in the image are too close to provide extra information for depth inference. In our experiments, to determine the number of virtual planes, we compute the mean depth sampling interval for a corresponding 0.5 pixel distance in the image.

For determining the local search range for depth residual around the current depth estimate for each pixel, we first project its 3D point into source views, find points that are two pixels away from the its projection along the epipolar line in both directions(see Fig.~\ref{fig:reproj} ``2 pixel length''), and back project those two points into 3D rays. The intersection of these two rays with the visual ray in the reference view determines the search range for depth refinement on current level.

\vspace{0.1cm}
\noindent{\bf Depth Map Estimator}
 Similar to MVSNet~\cite{yao2018mvsnet}, we apply 3D convolution to the constructed cost volume pyramid $\{{\bf C}^l\}_{l=0}^L$ to aggregate context information and output probability volumes $\{{\bf P}^l\}_{l=0}^L$, where ${\bf P}^l\in \R^{H/2^l\times W/2^l\times M}$. Detailed 3D convolution network design is in Supp. Mat. Note that ${\bf P}^L$ and $\{{\bf P}^l\}_{l=0}^{L-1}$ are generated on absolute and residual depth, respectively. We therefore first apply soft-argmax to ${\bf P}^L$ to obtain the coarse depth map. Then, we iteratively refine the obtained depth map by applying soft-argmax to $\{{\bf P}^l\}_{l=1}^{L-1}$ to obtain the depth residual for higher resolutions. 
 
Recall that sampled depth is $d =d_{\text{min}}+m (d_{\text{max}} - d_{\text{min}})/M, m\in\{0,1,2,\cdots,M-1\}$ at level $L$. Therefore, the depth estimate for each pixel ${\bf p}$\comment{at $L$th level} is computed as
\begin{equation}
\vspace{-0.2cm}
{\bf D}^L({\bf p}) = \sum_{m=0}^{M-1}d{\bf P}^L_{\bf p}(d).
\vspace{-0.1cm}
\end{equation}
To further refine the current estimate which is either the coarse depth map or a refined depth at ($l+1$)th level, we estimate the~\emph{residual depth}. Assume $r_{\bf p} = m\cdot \Delta d_{\bf p}^l$ denotes the depth residual hypothesis. We compute the updated depth at the next level as 
\begin{equation}
{\bf D}^l({\bf p}) = {\bf D}^{l+1}_{\uparrow} ({\bf p})+\sum_{m =-M/2}^{(M-2)/2}r_{\bf p}{\bf P}^l_{\bf p}(r_{\bf p})
\vspace{-0.1cm}
\end{equation}
 where $l\in\{L-1, L-2, \cdots, 0\}$. In our experiments, we observe no depth map refinement after our pyramidal depth estimation is further required to obtain good results.
 
\subsection{Loss Function}
We adopt a supervised learning strategy and construct the pyramid for ground truth depth $\{{\bf D}_{\text GT}^l\}_{l=0}^L$ as supervisory signal. Similar to existing MVSNet framework \cite{yao2018mvsnet}, we make use of the $l_1$ norm measuring the absolute difference between the ground truth and the estimated depth. For each training sample, our loss is
\begin{equation}
Loss=\sum_{l=0}^{L}\sum_{{\bf p}\in \Omega}\|{\bf D}_{\text GT}^l({\bf p})-{\bf D}^{l}({\bf p})\|_1,
\end{equation}
where $\Omega$ is the set of valid pixels with ground truth measurements.

\section{Experiments}
In this section, we demonstrate the performance of our framework for MVS with a comprehensive set of experiments in standard benchmarks. Below, we first describe the datasets and benchmarks and then analyze our results.
\subsection{Datasets}
 \noindent\textbf{DTU Dataset}~\cite{aanaes2016large} is a large-scale MVS dataset with 124 scenes scanned from 49 or 64 views under 7 different lighting conditions. DTU provides 3D point clouds acquired using structured-light sensors.~Each view consists of an image and the calibrated camera parameters.~To train our model, we generate a $160\times 128$ depth map for each view by using the method provided by MVSNet~\cite{yao2018mvsnet}. We use the same training, validation and evaluation sets as defined in~\cite{yao2018mvsnet,yao2019recurrent}.
 
 \noindent\textbf{Tanks and Temples}~\cite{knapitsch2017tanks} contains both indoor and outdoor scenes under realistic lighting conditions with large scale variations. For comparison with other approaches, we evaluate our results on the \emph{intermediate set}.
\subsection{Implementation}
\noindent\textbf{Training} We train our CVP-MVSNet on DTU training set.  
 Unlike previous methods~\cite{yao2018mvsnet,yao2019recurrent} that take high resolution image as input but estimate a depth map of smaller size, our method produces the same size depth map as the input image.
 For training, we match the ground-truth depth map by downsampling the high resolution image into a smaller one of size $160\times 128$.~Then, we build the image and ground truth depth pyramid with 2 levels.~To construct the~\emph{cost volume pyramid}, we uniformly sample $M=48$ depth hypotheses across entire depth range at the coarsest ($2$nd) level.~Then, each pixel has $M=8$ depth residual hypotheses at the next level for the refinement of the depth estimation.~Following MVSNet~\cite{yao2018mvsnet}, we adopt 3 views for training. We implemented our network using Pytorch~\cite{paszke2017automatic}, and we used ADAM~\cite{kingma2014adam} to train our model. The batch size is set to 16 and the network is end-to-end trained on a NVIDIA TITAN RTX graphics card for 27 epochs. The learning rate is initially set to 0.001 and divided by 2 iteratively at the $10^{th}$,$12^{th}$,$14^{th}$ and $20^{th}$ epoch.

\noindent\textbf{Metrics}. We follow the standard evaluation protocol as in~\cite{aanaes2016large,yao2018mvsnet}. We report the~\emph{accuracy},~\emph{completeness} and~\emph{overall score} of the reconstructed point clouds. \emph{Accuracy} is measured as the distance from  estimated point clouds to the ground truth ones in millimeter and~\emph{completeness} is defined as the distance from ground truth point clouds to the estimated ones~\cite{aanaes2016large}. The~\emph{overall score} is the average of accuracy and completeness~\cite{yao2018mvsnet}. 

\vspace{0.1cm}
\noindent\textbf{Evaluation} As the parameters are shared across the~\emph{cost volume pyramid}, we can evaluate our model with different number of cost volumes and input views.~For the evaluation, we set the number of depth sampling, $M=96$ for the coarsest depth estimation (same as~\cite{chen2019point}. We also provide results of $M=48$ in the Supp. Mat.) and $M=8$ for the following depth residual inference levels. Similar to previous methods~\cite{chen2019point,yao2018mvsnet,yao2019recurrent}, we use 5 views and apply the same depth map fusion method to obtain the point clouds. We evaluate our model with images of different size and set the pyramid levels accordingly to maintain a similar size as the input image ($80\times 64$) at coarsest level. For instance, for an input size of $1600\times1184$, the pyramid has 5 levels and 4 levels for an input size of $800\times576$ and $640\times 480$.

\begin{table}[!t]
\begin{center}
\footnotesize
\begin{tabular}{ll|ccc}
\hline
\multicolumn{2}{c|}{Method} & Acc. & Comp. & Overall (\textit{mm}) \\
\hline\hline
\parbox[t]{2mm}{\multirow{5}{*}{\rotatebox[origin=c]{90}{Geometic}}}
& Furu\cite{furu2010} & 0.613 & 0.941 & 0.777 \\
&Tola\cite{tola2012} & 0.342 & 1.190 & 0.766 \\
&Camp\cite{comp2008} & 0.835 & 0.554 & 0.695 \\
&Gipuma\cite{galliani2016gipuma} & \textbf{0.283} & 0.873 & 0.578 \\
&Colmap\cite{schoenberger2016sfm,schoenberger2016mvs} & 0.400 & 0.664 & 0.532 \\
\hline
\parbox[t]{2mm}{\multirow{5}{*}{\rotatebox[origin=c]{90}{Learning}}}
&SurfaceNet\cite{ji2017surfacenet} & 0.450 & 1.040 & 0.745 \\
&MVSNet\cite{yao2018mvsnet} & 0.396 & 0.527 & 0.462 \\
&P-MVSNet\cite{luo2019p} & 0.406 & 0.434 & 0.420 \\
&R-MVSNet\cite{yao2019recurrent} & 0.383 & 0.452 & 0.417 \\
&MVSCRF\cite{xue2019mvscrf} & 0.371 & 0.426 & 0.398 \\
&Point-MVSNet\cite{chen2019point} & 0.342 & \underline{0.411} & \underline{0.376} \\\hline
\multicolumn{2}{c|}{Ours} & \underline{0.296} & \textbf{0.406} & \textbf{0.351} \\
\hline
\end{tabular}
\end{center}
\vspace{-0.3cm}
\caption{Quantitative results of reconstruction quality on DTU dataset (lower is better). Our method outperforms all methods on Mean Completeness and Overall reconstruction quality and achieved second best on Mean Accuracy.
}
\label{table:quatitative}
\vspace{-0.4cm}
\end{table}
\subsection{Results on DTU dataset}

We first compare our results to those reported by traditional geometric-based methods and other learning-based baseline methods. As summarized in Table~\ref{table:quatitative}, our method outperforms all current learning-based methods in terms of~\emph{accuracy},~\emph{completeness} and~\emph{overall score}. Compared to geometric-based approaches, only the method proposed by Galliani \etal~\cite{galliani2016gipuma} provides slightly better results in terms of mean~\emph{accuracy}.

We now compare our results to related learning based methods in terms of GPU memory usage and runtime for different input resolution.~The summary of these results is listed in Table \ref{table:dtu-speed}.~As shown, our network, with a similar memory usage (bottom row), is able to produce better point clouds with \comment{significantly} lower runtime. In addition, compared to Point-MVSNet~\cite{chen2019point} on the same size of depth map output (top rows), our approach is six times faster and consumes six times less memory \comment{provides} with similar accuracy. We can output high resolution depth map with better accuracy, less memory usage and shorter runtime than Point-MVSNet~\cite{chen2019point}.

  \begin{figure*}[!t]
    \begin{center}
    \setlength\tabcolsep{1pt}
    \begin{tabular}{cccc}
          \includegraphics[width=0.225\linewidth]{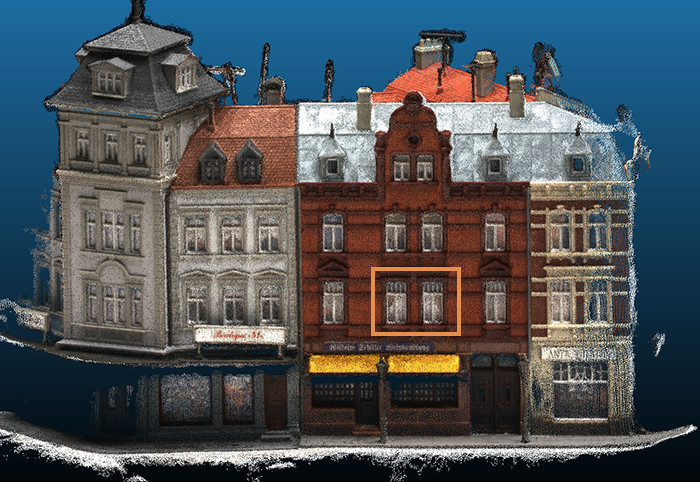}
          & \includegraphics[width=0.225\linewidth]{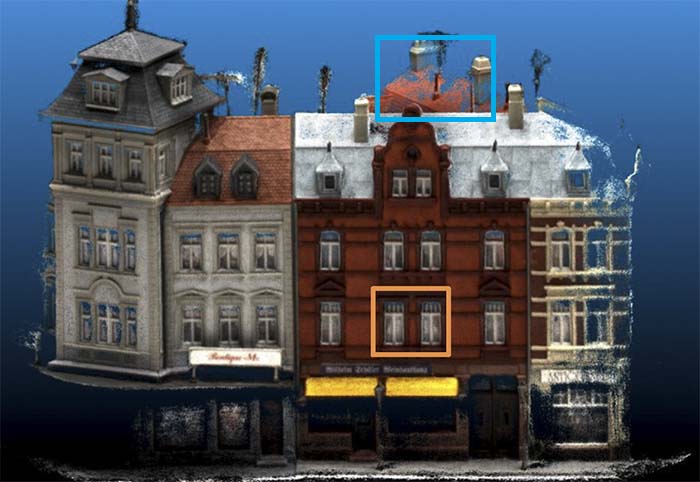}
          & \includegraphics[width=0.225\linewidth]{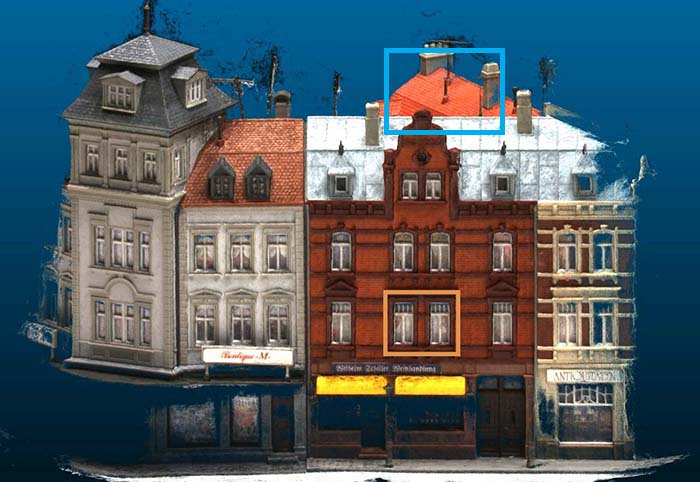}
          & \includegraphics[width=0.225\linewidth]{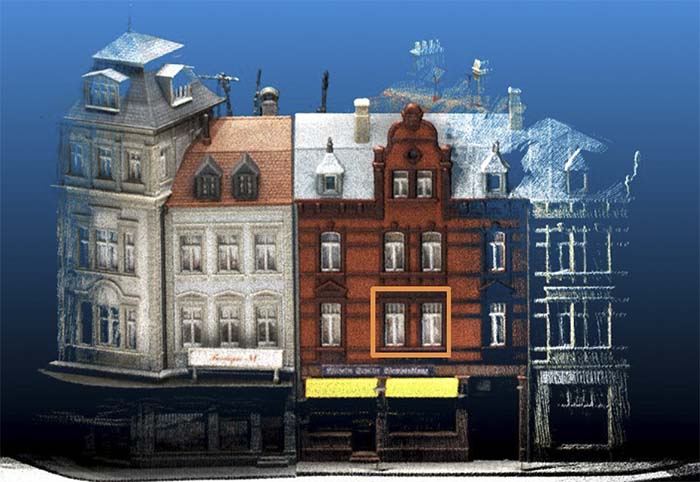}\\
          \includegraphics[width=0.225\linewidth]{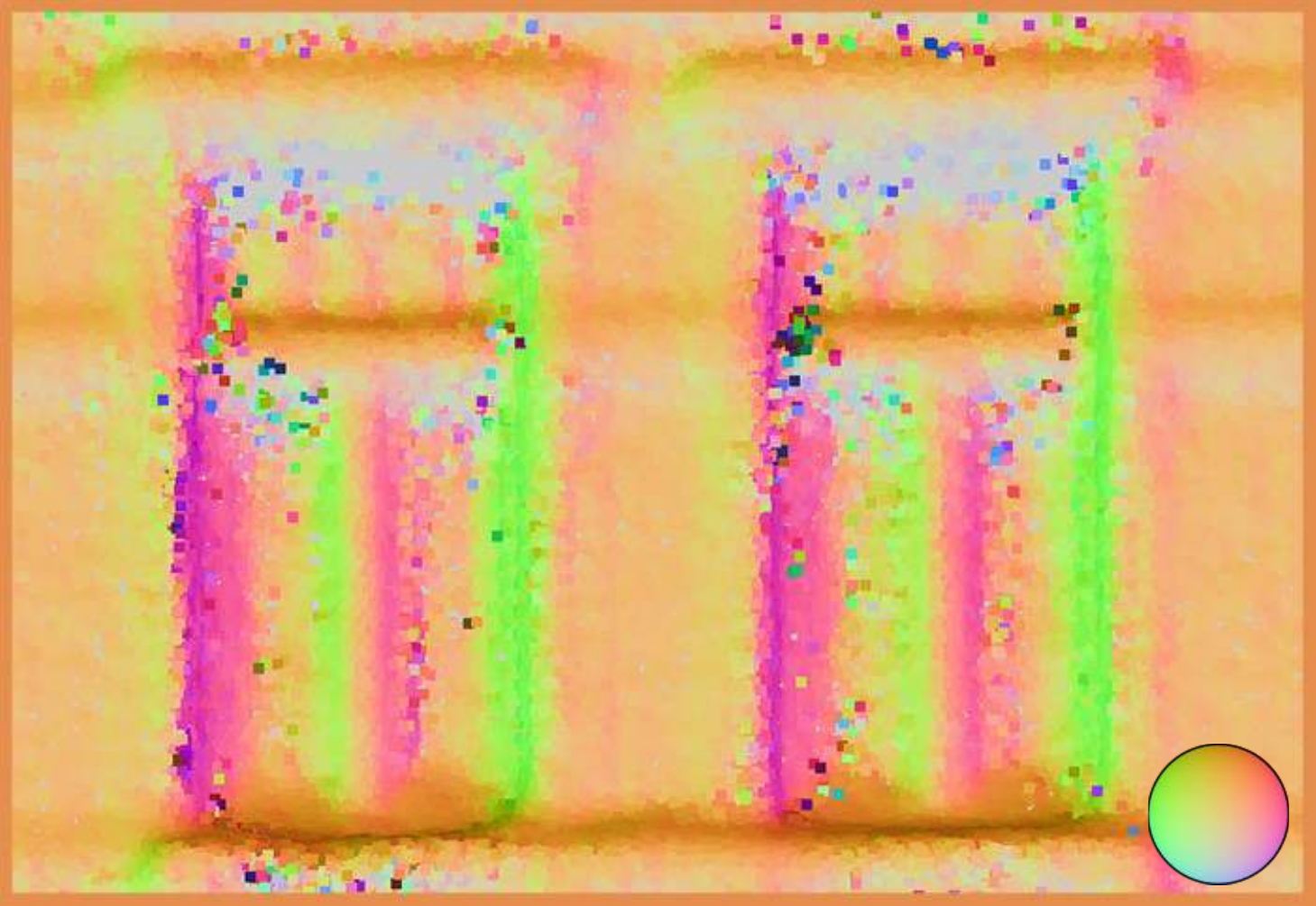}
          & \includegraphics[width=0.225\linewidth]{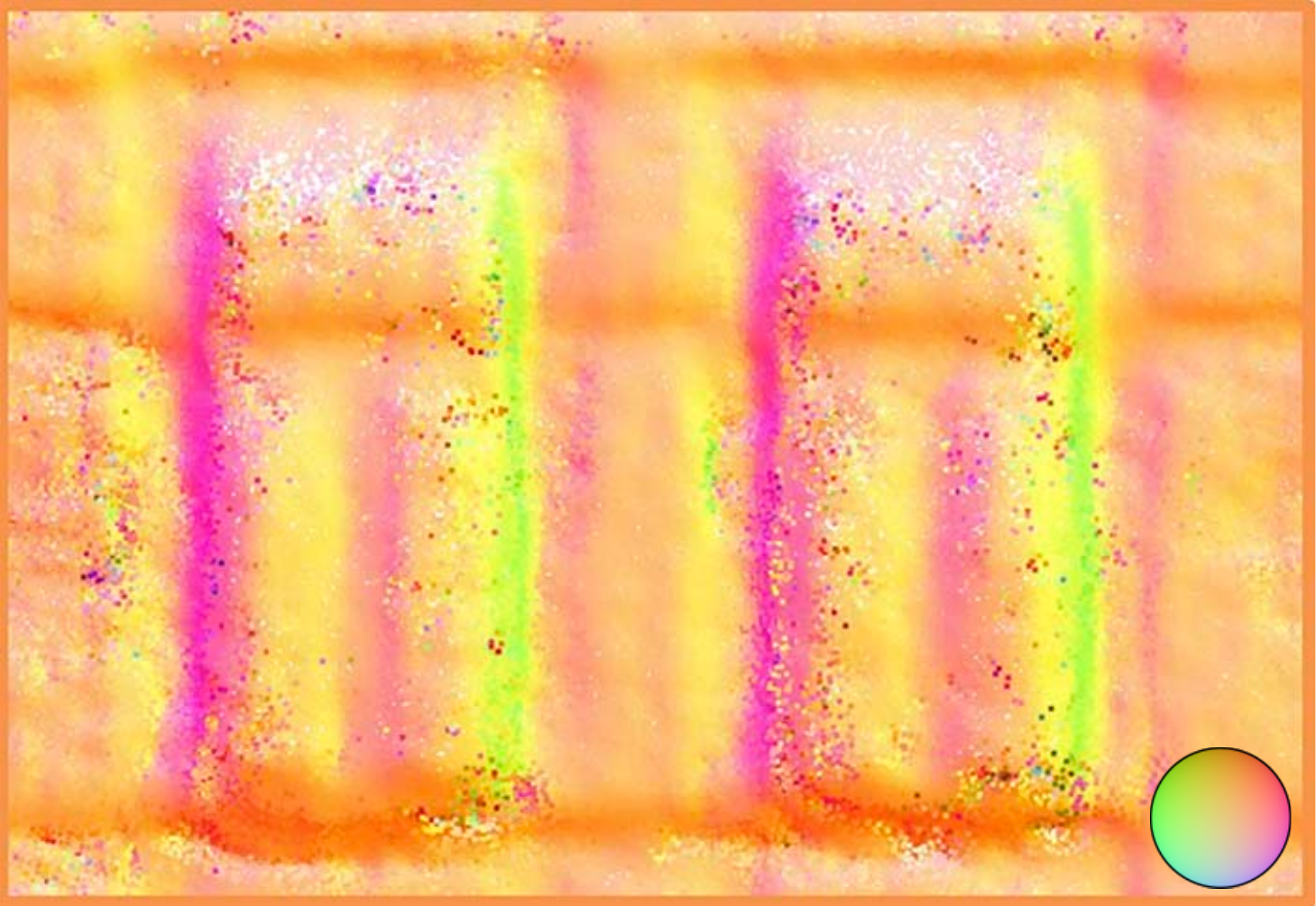}
          & \includegraphics[width=0.225\linewidth]{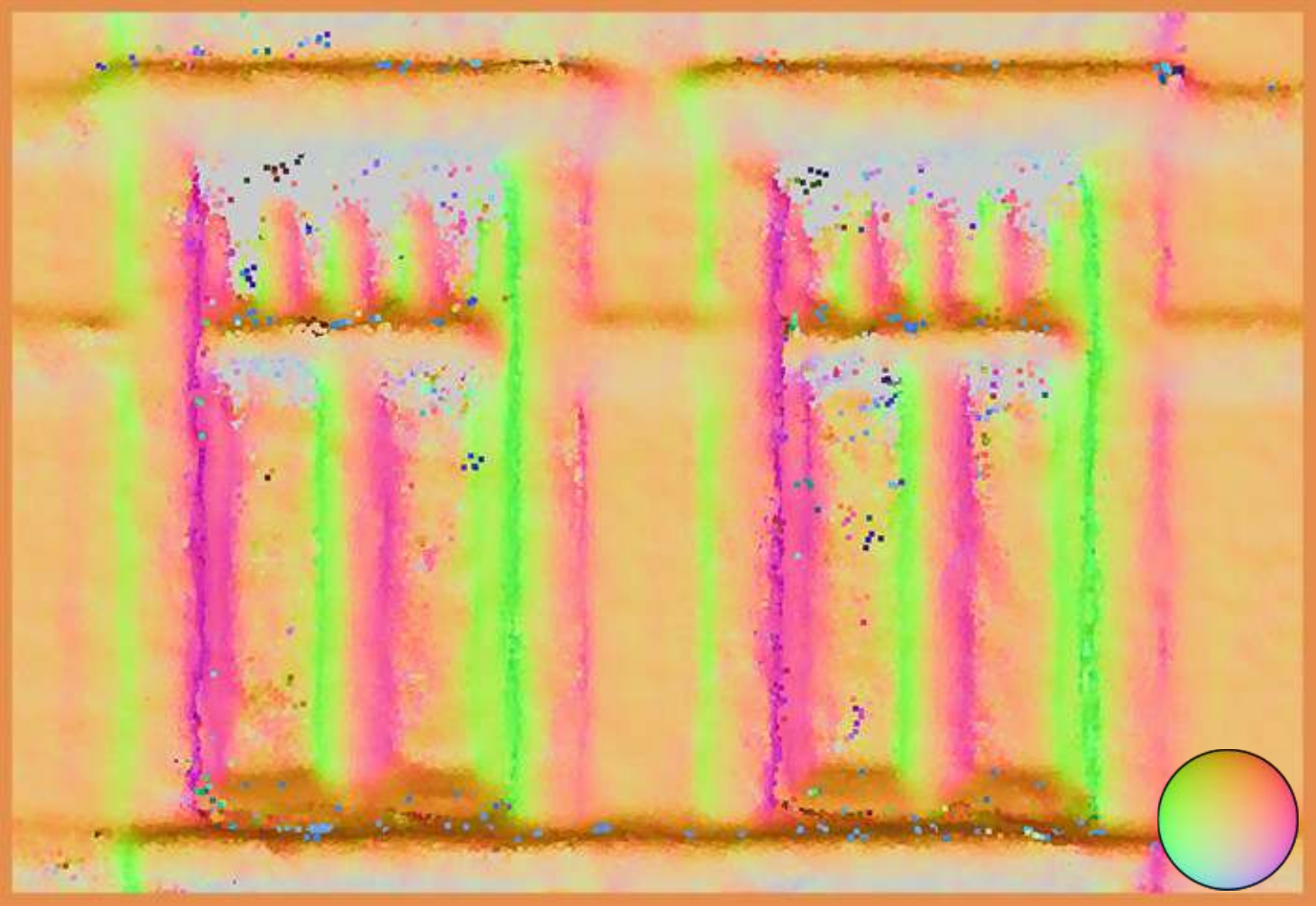}
          & \includegraphics[width=0.225\linewidth]{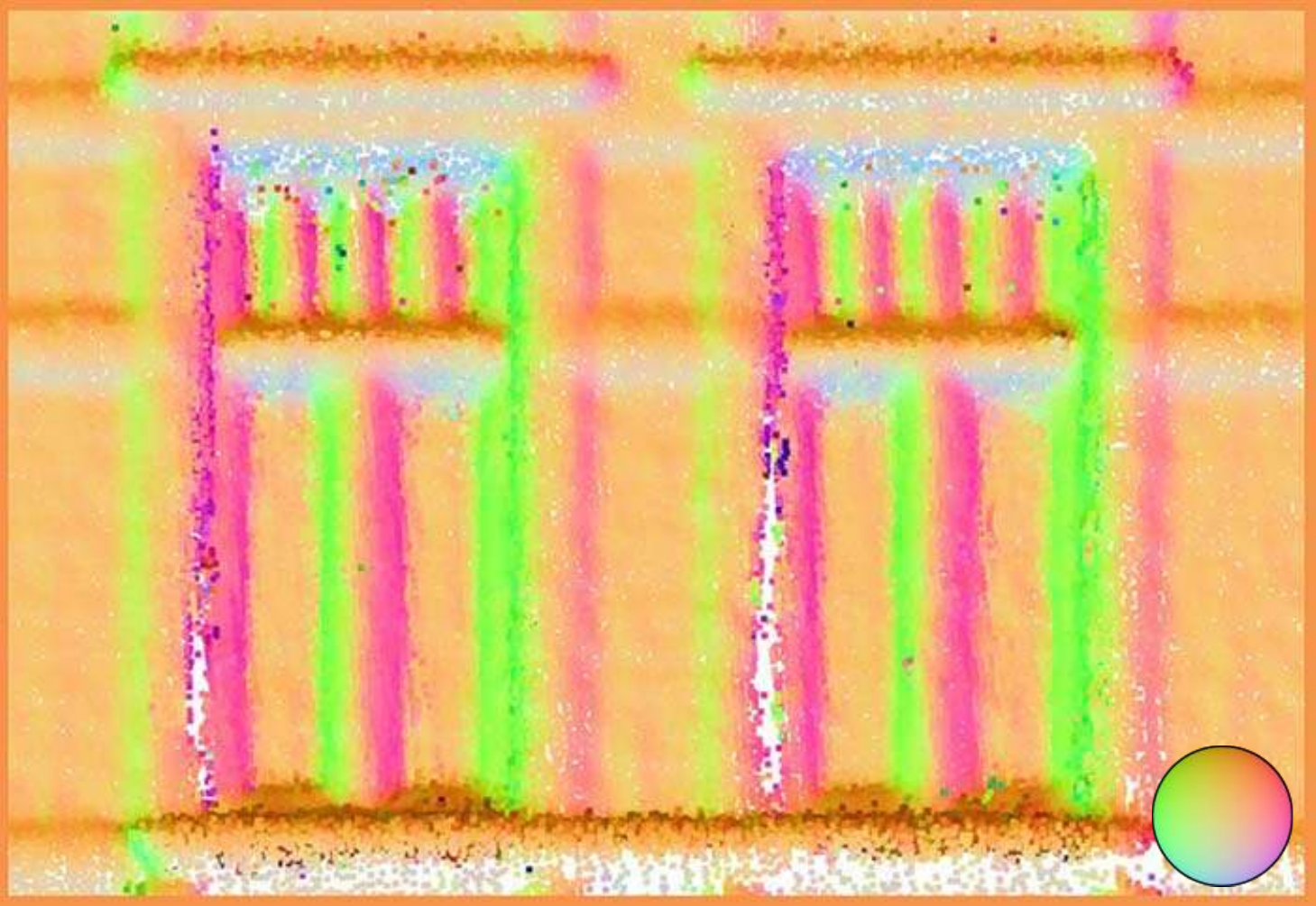}\\
          R-MVSNet~\cite{yao2019recurrent} & Point-MVSNet~\cite{chen2019point} & Ours & Ground truth
    \end{tabular}
    \end{center}
    \vspace{-0.5cm}
    \caption{Qualitative results of scan 9 of DTU dataset. The upper row shows the point clouds and the bottom row shows the normal map corresponding to the orange rectangle. As highlighted in the blue rectangle, the completeness of our results is better than those provided by Point-MVSNet\cite{chen2019point}. The normal map (orange rectangle) further shows that our results are smoother on surfaces while maintaining more high-frequency details. 
    }
    \label{fig:normDTU}
    \vspace{-0.1cm}
\end{figure*}
 
\begin{table*}[!t]
\begin{center}
\resizebox{0.925\linewidth}{!}{
\footnotesize
\begin{tabular}{l|cccccccc}
\hline
Method       & Input Size & Depth Map Size & Acc.\scriptsize(mm) & Comp.\scriptsize{(mm)} & Overall\scriptsize{(mm)} & \textit{f-score}\scriptsize{(0.5mm)} & GPU Mem\scriptsize{(MB)} & Runtime\scriptsize{(s)} \\
\hline\hline
Point-MVSNet\cite{chen2019point} & 1280x960 & 640x480        & 0.361     & 0.421      & 0.391 & 84.27       & 8989         & 2.03            \\
Ours-640     & 640x480    & 640x480         & 0.372     & 0.434      & 0.403       & 82.44 & \textbf{1416}         & \textbf{0.37}        \\
\hline
Point-MVSNet\cite{chen2019point} & 1600x1152  & 800x576        & 0.342     & 0.411      & 0.376  & -      & 13081             & 3.04            \\
Ours-800     & 800x576    & 800x576         & 0.340    & 0.418     & 0.379    & 86.82    & \textbf{2207}         & \textbf{0.49}        \\
\hline
MVSNet\cite{yao2018mvsnet}       & 1600x1152  & 400x288        & 0.396     & 0.527      & 0.462   & 78.10     & 22511        & 2.76            \\
R-MVSNet\cite{yao2019recurrent}     & 1600x1152  & 400x288        & 0.383     & 0.452      & 0.417  & 83.96      & \textbf{6915}         & 5.09            \\
Point-MVSNet\cite{chen2019point} & 1600x1152  & 800x576        & 0.342     & 0.411      & 0.376  & -      & 13081             & 3.04            \\
Ours         & 1600x1152  & 1600x1152      & \textbf{0.296}     & \textbf{0.406}     & \textbf{0.351}  & \textbf{88.61}     & 8795         & \textbf{1.72}        \\
\hline
\end{tabular}
}
\end{center}
\vspace{-0.5cm}
\caption{Comparison of reconstruction quality, GPU memory usage and runtime on DTU dataset for different input sizes. GPU memory usage and runtime are obtained by running the official evaluation code of baselines on a same machine with a NVIDIA TITAN RTX graphics card. For the same size of depth maps (Ours-640, Ours-800) and a performance similar to Point-MVSNet~\cite{chen2019point}, our method is 6 times faster and consumes 6 times smaller GPU memory. For the same size of input images (Ours), our method achieves the best reconstruction with the shortest time and a reasonable GPU memory usage. 
}

\label{table:dtu-speed}
\vspace{-0.14cm}
\end{table*}

\begin{table*}[!t]
\begin{center}
\footnotesize
\begin{tabular}{l|cccccccccc}
\hline
Method       & Rank       & Mean  & Family & Francis & Horse & Lighthouse & M60   & Panther & Playground & Train \\
\hline\hline
P-MVSNet~\cite{luo2019p} & \textbf{11.72} & \textbf{55.62} & \underline{70.04} & \underline{44.64} & \textbf{40.22} & \textbf{65.20} & \underline{55.08} & \textbf{55.17} & \textbf{60.37} & \textbf{54.29} \\
Ours         & 12.75 & \underline{54.03} & \textbf{76.5}   & \textbf{47.74}   & \underline{36.34} & \underline{55.12}      & \textbf{57.28} & \underline{54.28}   & \underline{57.43}      & \underline{47.54} \\
Point-MVSNet\cite{chen2019point} & 29.25      & 48.27 & 61.79  & 41.15   & 34.20  & 50.79      & 51.97 & 50.85   & 52.38      & 43.06 \\
R-MVSNet\cite{yao2019recurrent}     & 31.75      & 48.40  & 69.96  & 46.65   & 32.59 & 42.95      & 51.88 & 48.80    & 52.00         & 42.38 \\
MVSNet\cite{yao2018mvsnet}       & 42.75      & 43.48 & 55.99  & 28.55   & 25.07 & 50.79      & 53.96 & 50.86   & 47.90       & 34.69 \\
\hline
\end{tabular}
\end{center}
\vspace{-0.4cm}
\caption{Performance on Tanks and Temples~\cite{knapitsch2017tanks} on November 12, 2019. Our results outperform Point-MVSNet~\cite{chen2019point}, which is the strongest baseline on DTU dataset, and are competitive compared to P-MVSNet~\cite{luo2019p}.}
\label{tanks}
\vspace{-0.4cm}
\end{table*}

\begin{figure*}[!ht]
    \begin{center}
    \setlength\tabcolsep{3pt}
    \begin{tabular}{cccc}
          \includegraphics[width=0.18\linewidth]{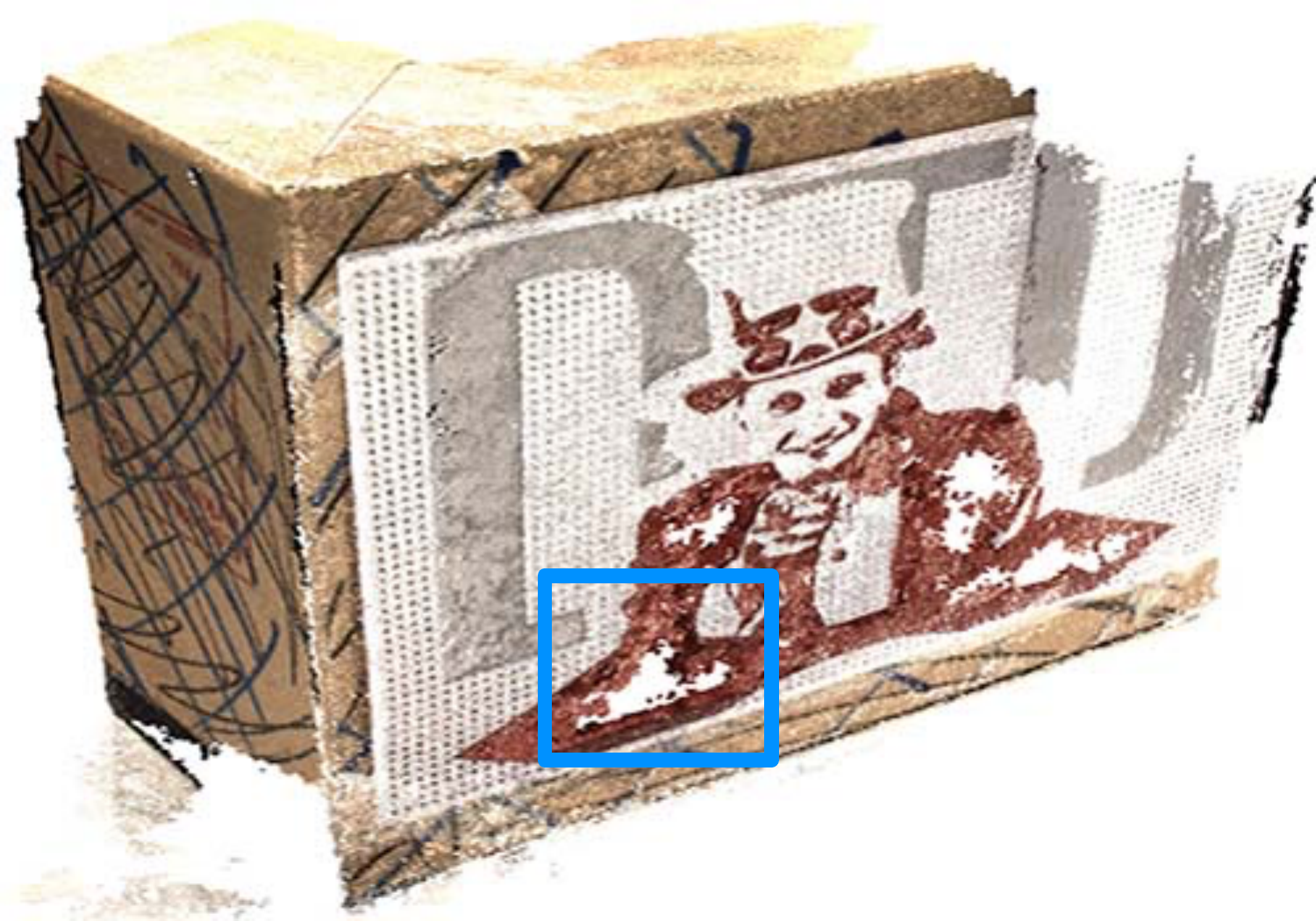}
          & \includegraphics[width=0.18\linewidth]{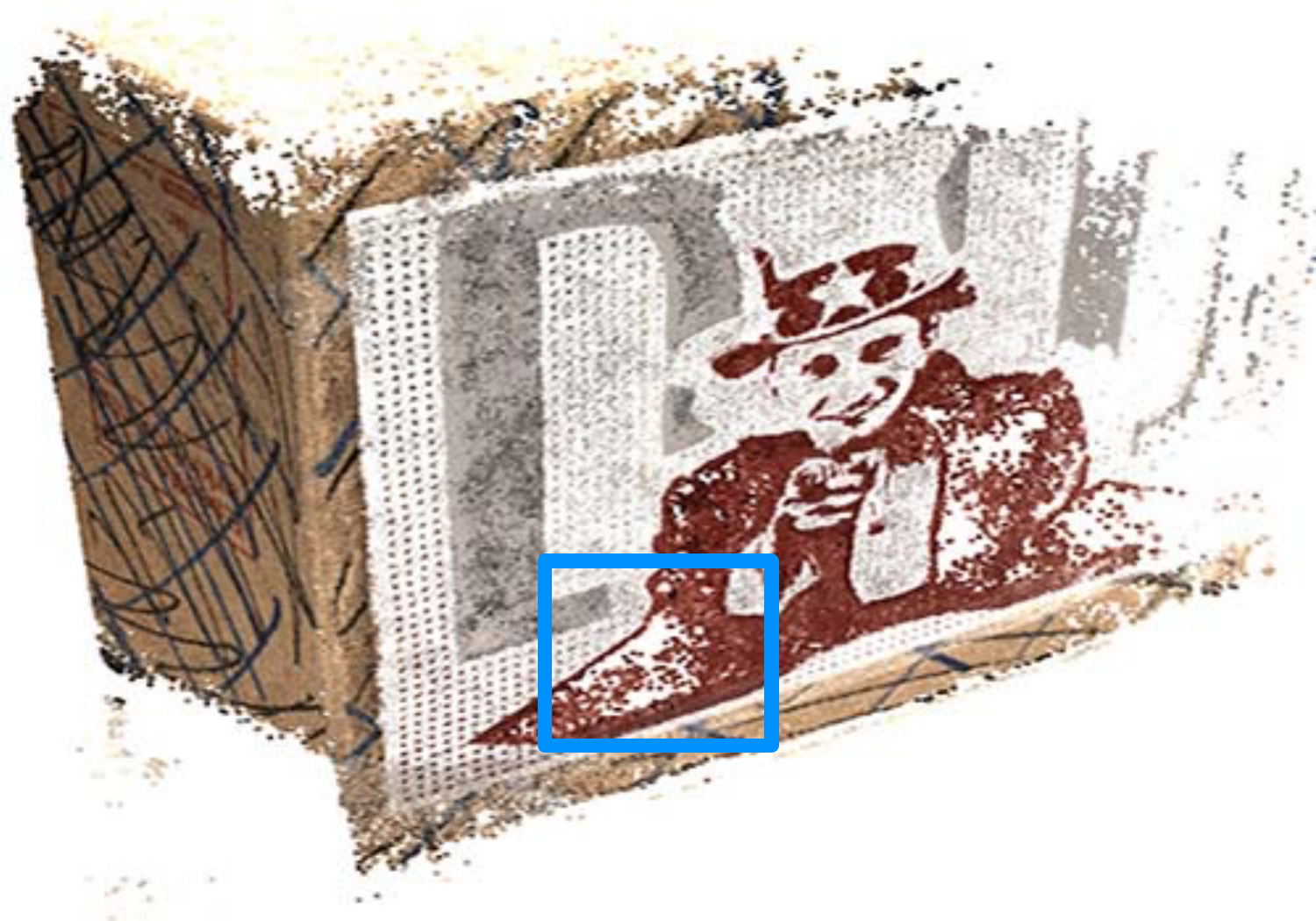}
          & \includegraphics[width=0.18\linewidth]{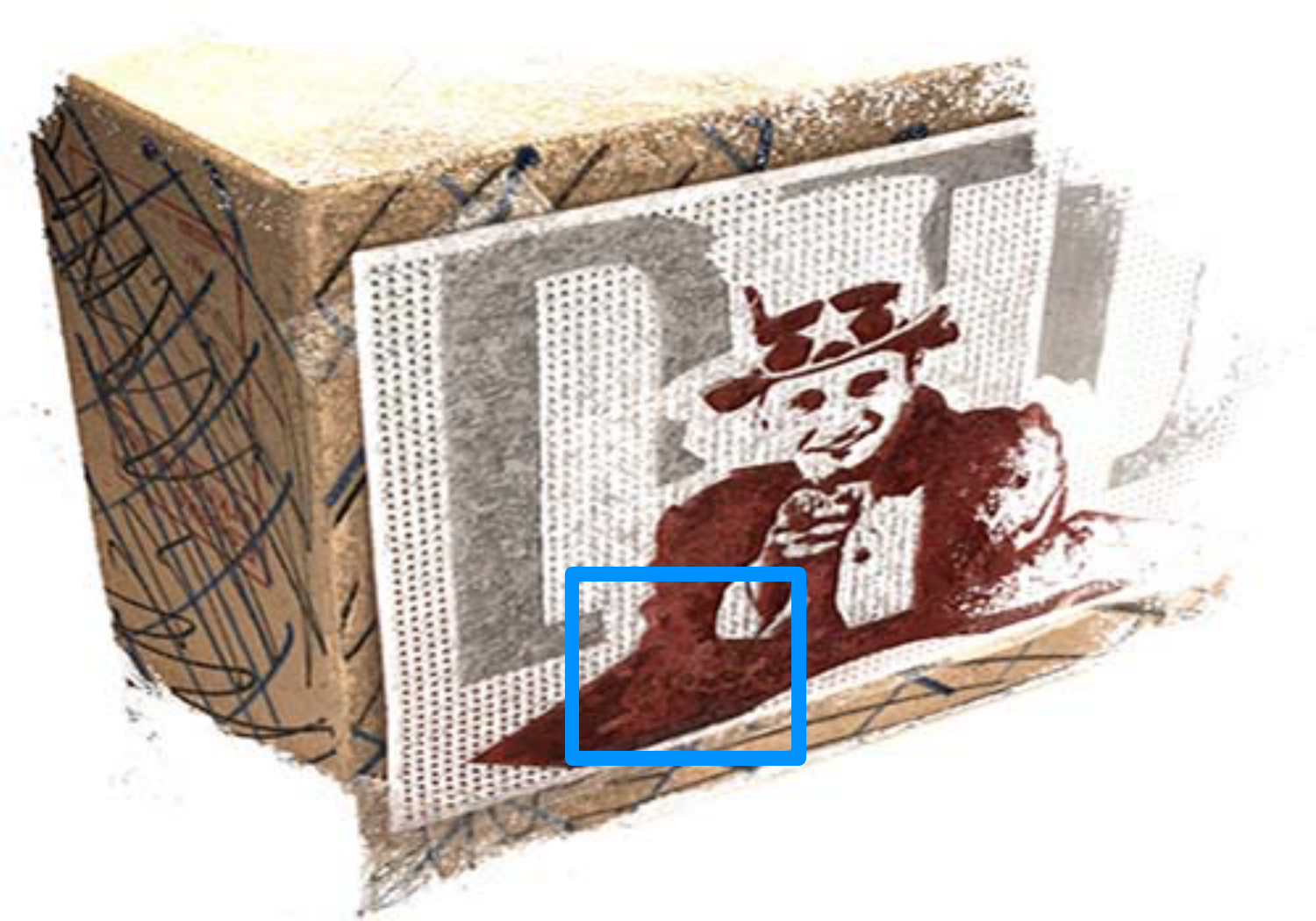}
          & \includegraphics[width=0.18\linewidth]{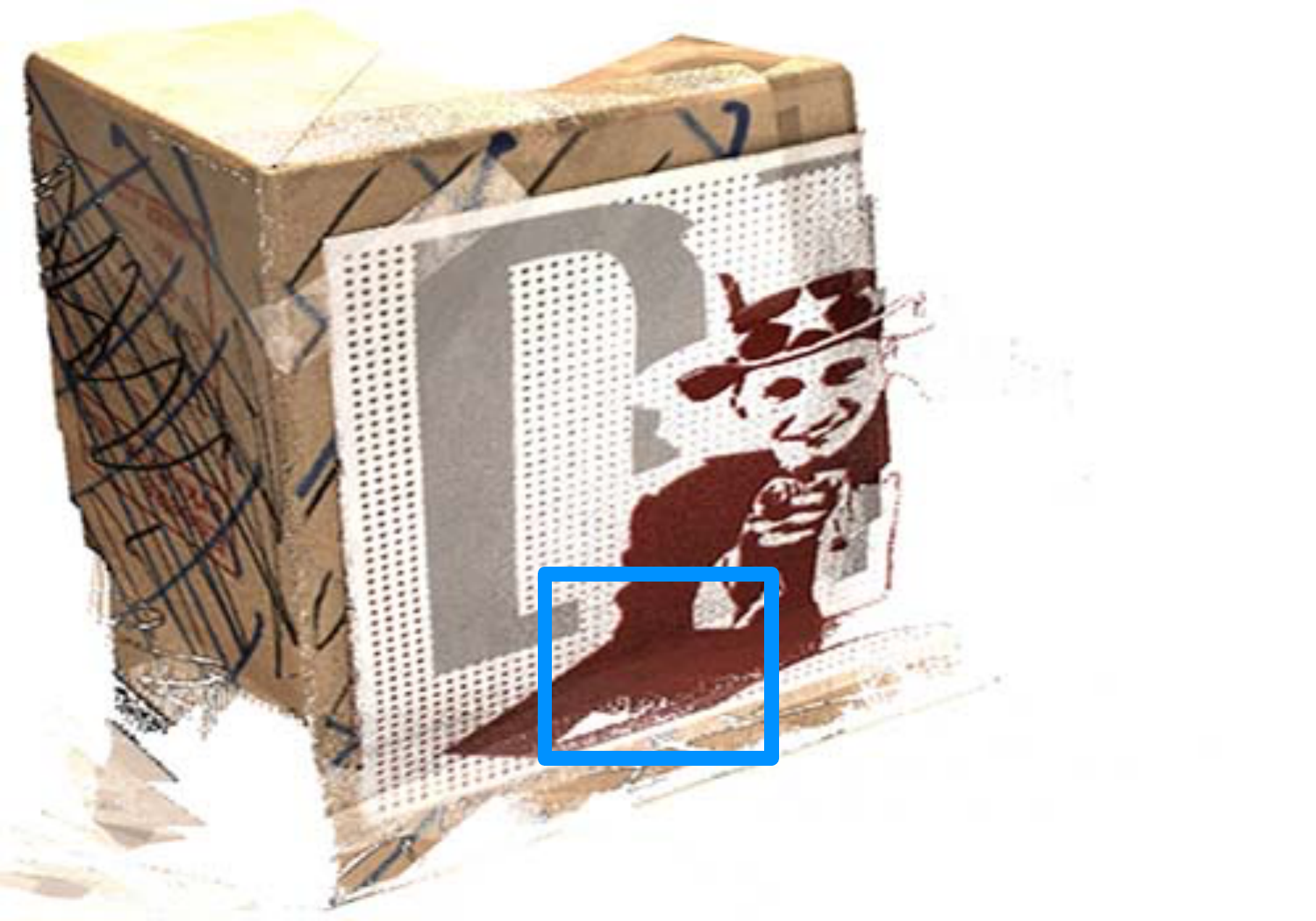}\\
          \includegraphics[width=0.18\linewidth]{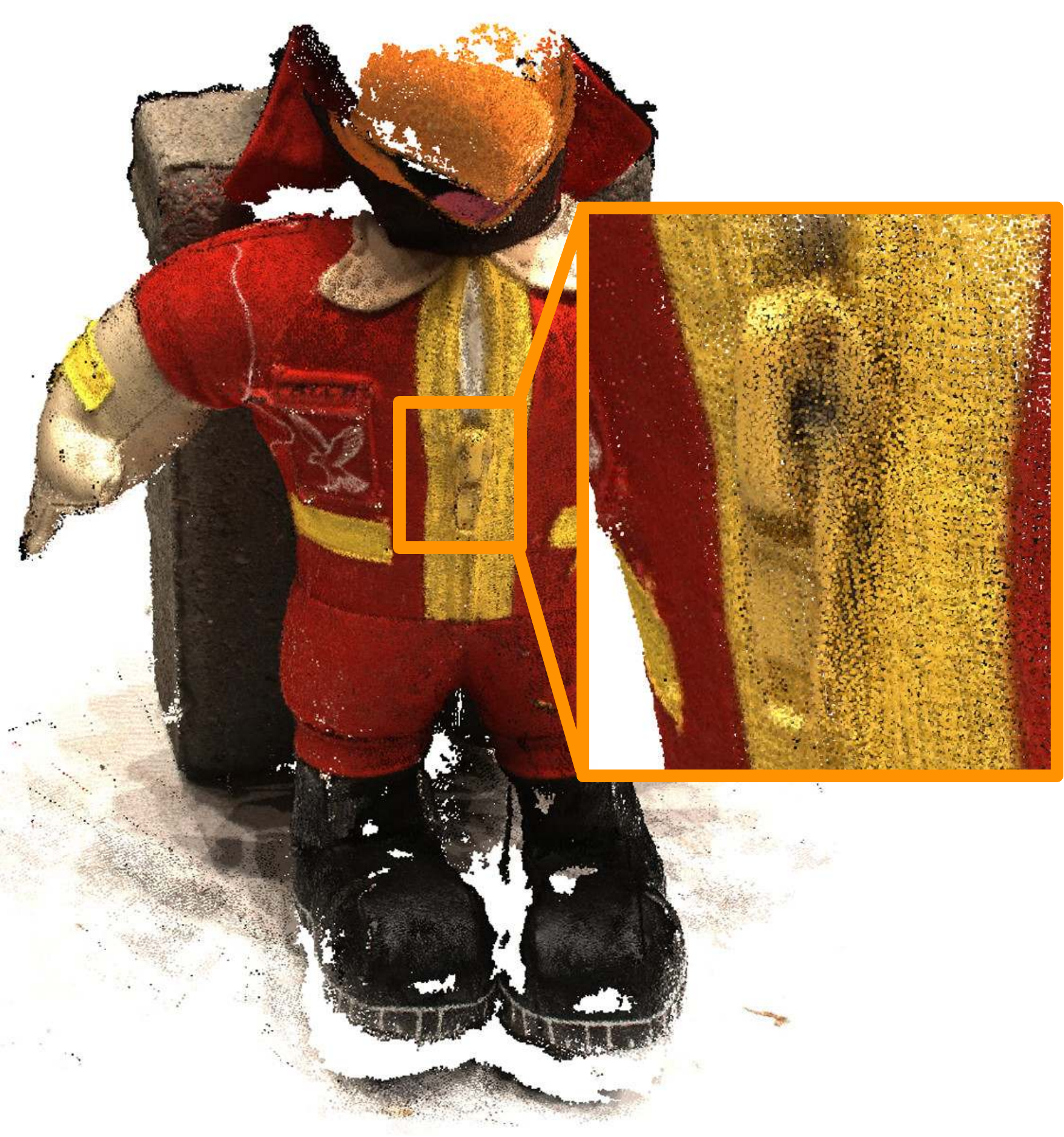}
          & \includegraphics[width=0.18\linewidth]{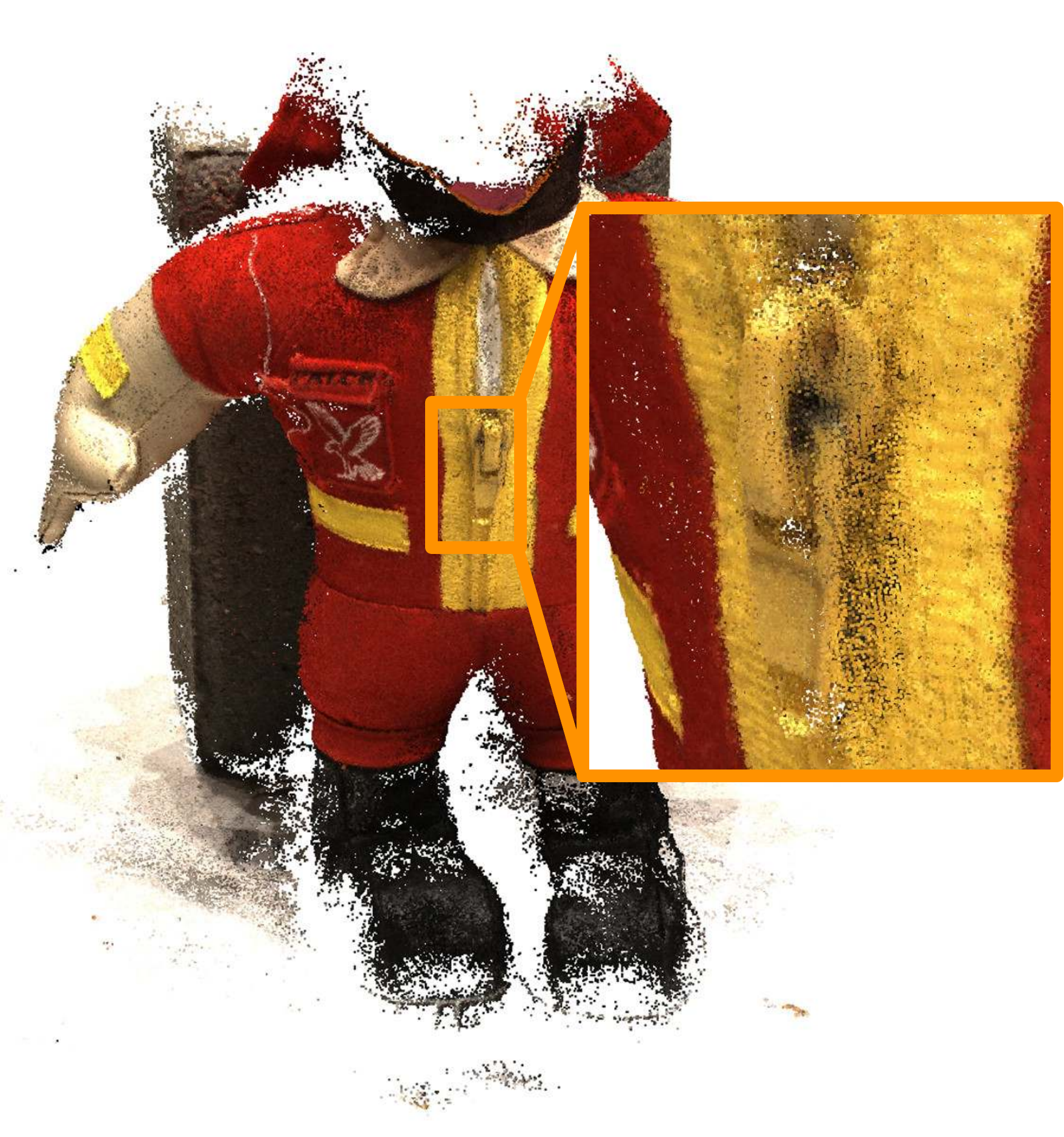}
          & \includegraphics[width=0.18\linewidth]{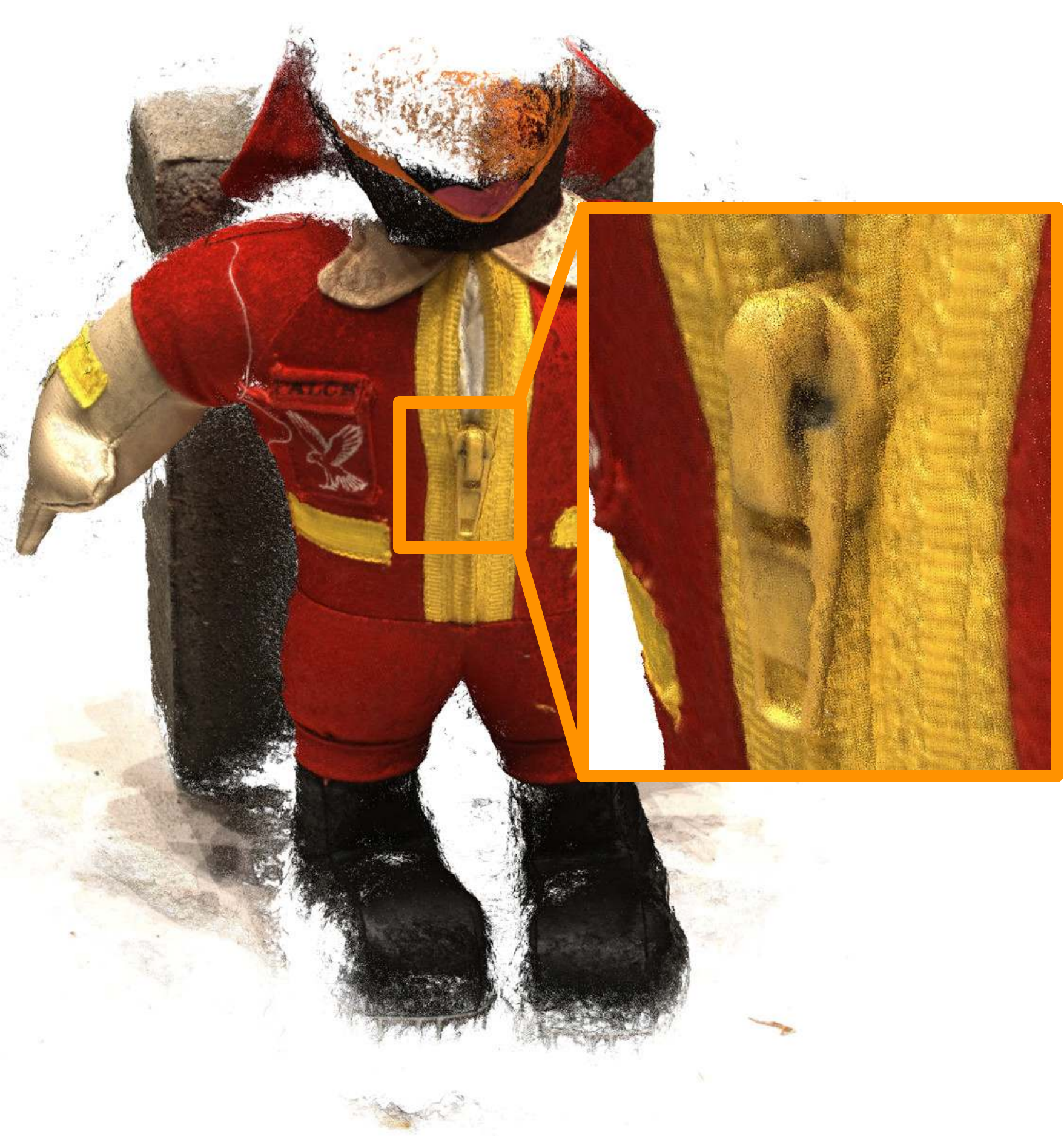}
          & \includegraphics[width=0.18\linewidth]{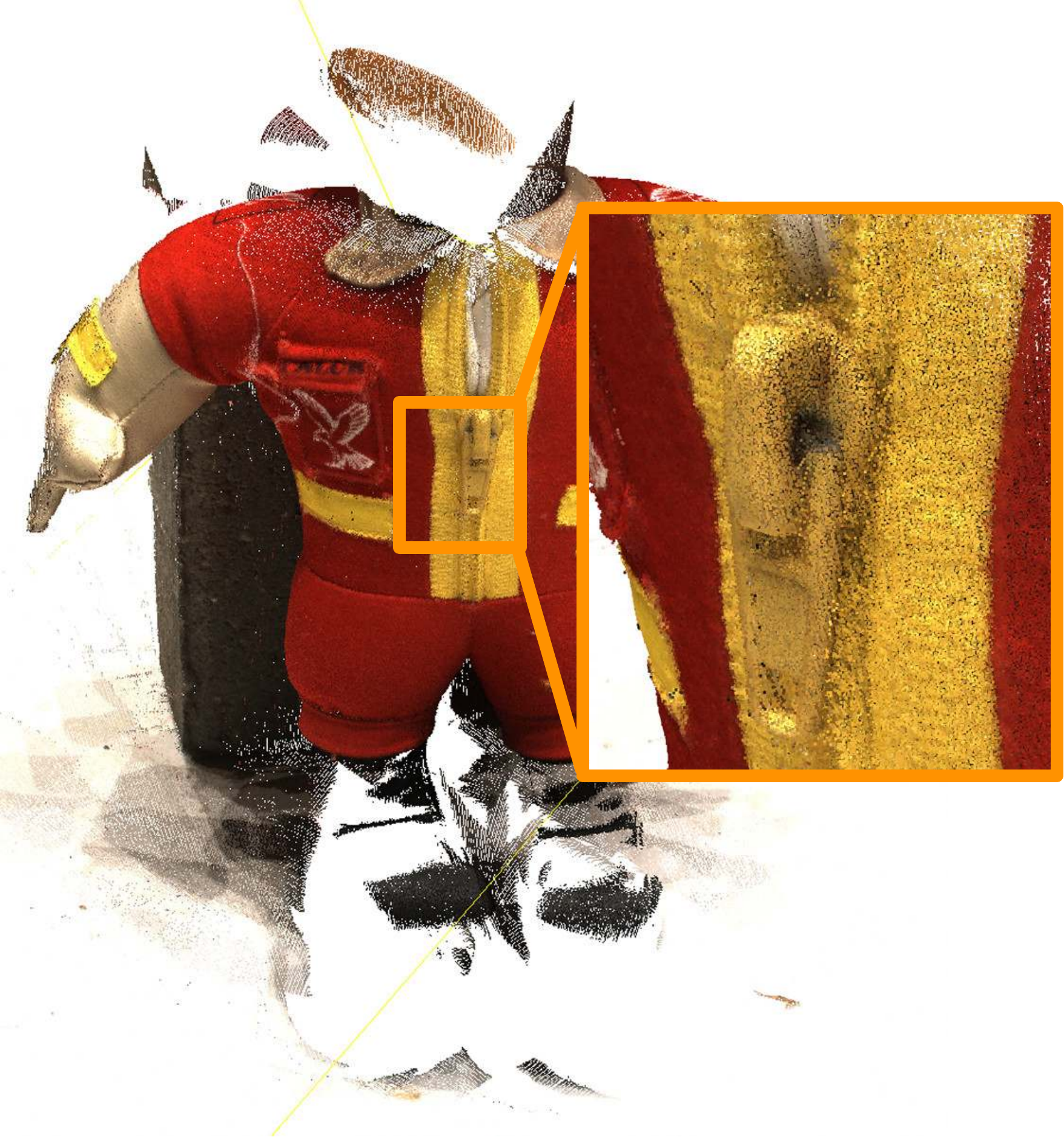}\\
          R-MVSNet~\cite{yao2019recurrent} & Point-MVSNet~\cite{chen2019point} & Ours & Ground truth
    \end{tabular}
    \end{center}
    \vspace{-0.5cm}
    \caption{Additional results from DTU dataset. Best viewed on screen.}
    \label{fig:qualityDTU}
    \vspace{-0.4cm}
\end{figure*}

Figures~\ref{fig:normDTU} and~\ref{fig:qualityDTU} show some qualitative results. As shown, our method is able to reconstruct more details than Point-MVSNet~\cite{chen2019point}, see for instance, the details highlighted in blue box of the roof behind the front building. Compared to R-MVSNet~\cite{yao2019recurrent} and Point-MVSNet~\cite{chen2019point}, as we can see in the normal maps, our results are smoother on the surfaces while capturing more high-frequency details in edgy areas. 
\begin{figure*}[!ht]
    \begin{center}
    \resizebox{0.7\linewidth}{!}{
    \setlength\tabcolsep{2pt}
    \begin{tabular}{cc}
          \includegraphics[width=0.405\linewidth]{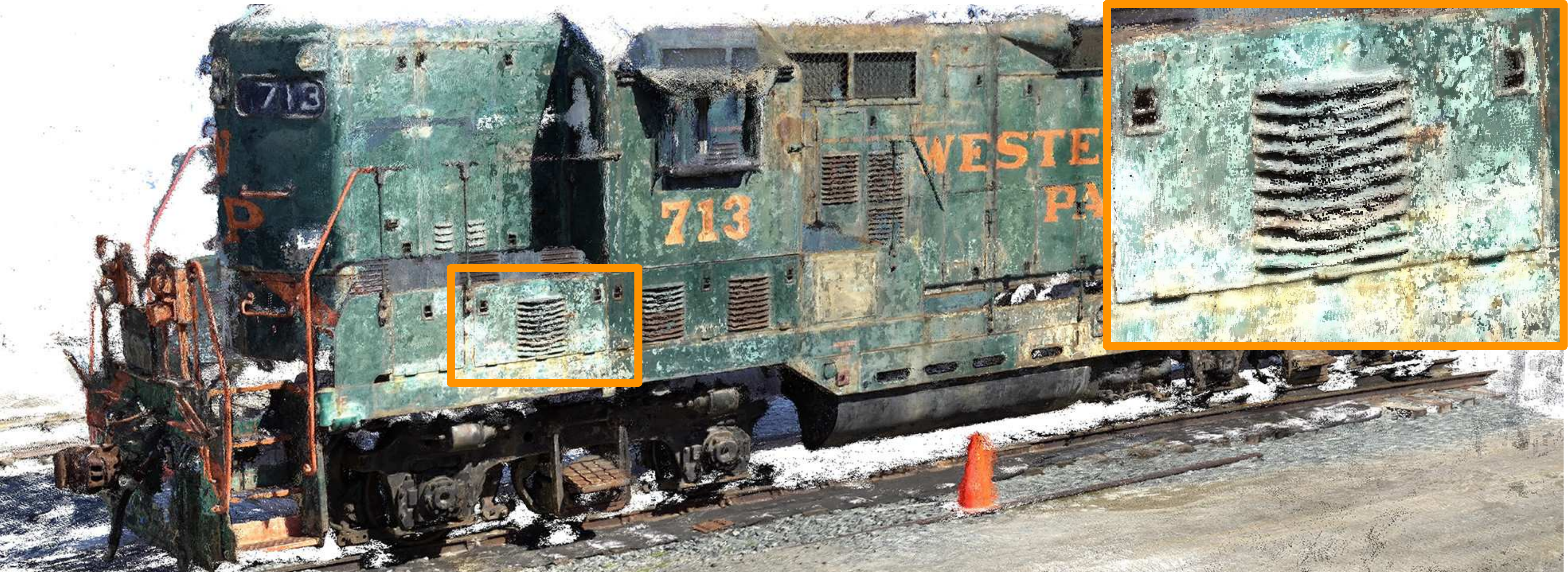}
          & \includegraphics[width=0.3650\linewidth]{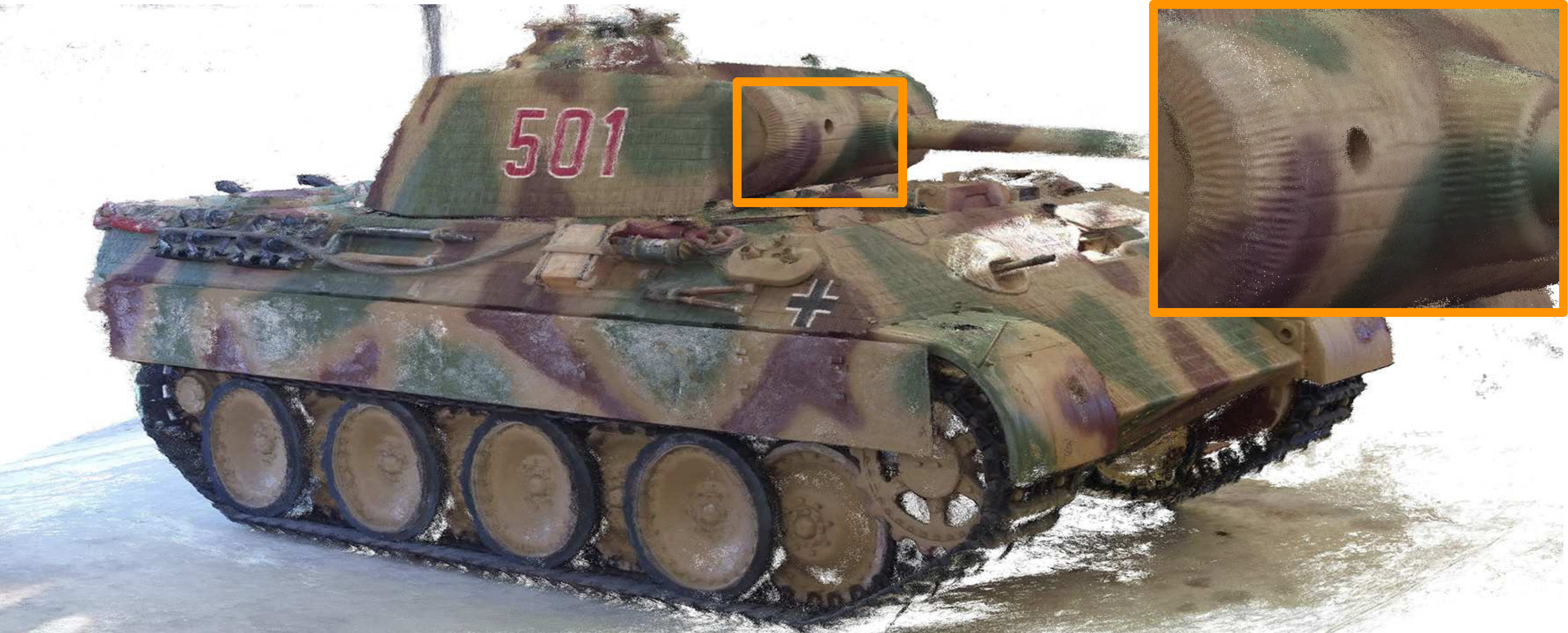}\\
         (a) Train & (b) Panther
    \end{tabular}
    }
    \resizebox{0.781\linewidth}{!}{
    \setlength\tabcolsep{1pt}
    \begin{tabular}{ccc}
          \includegraphics[width=0.284\linewidth]{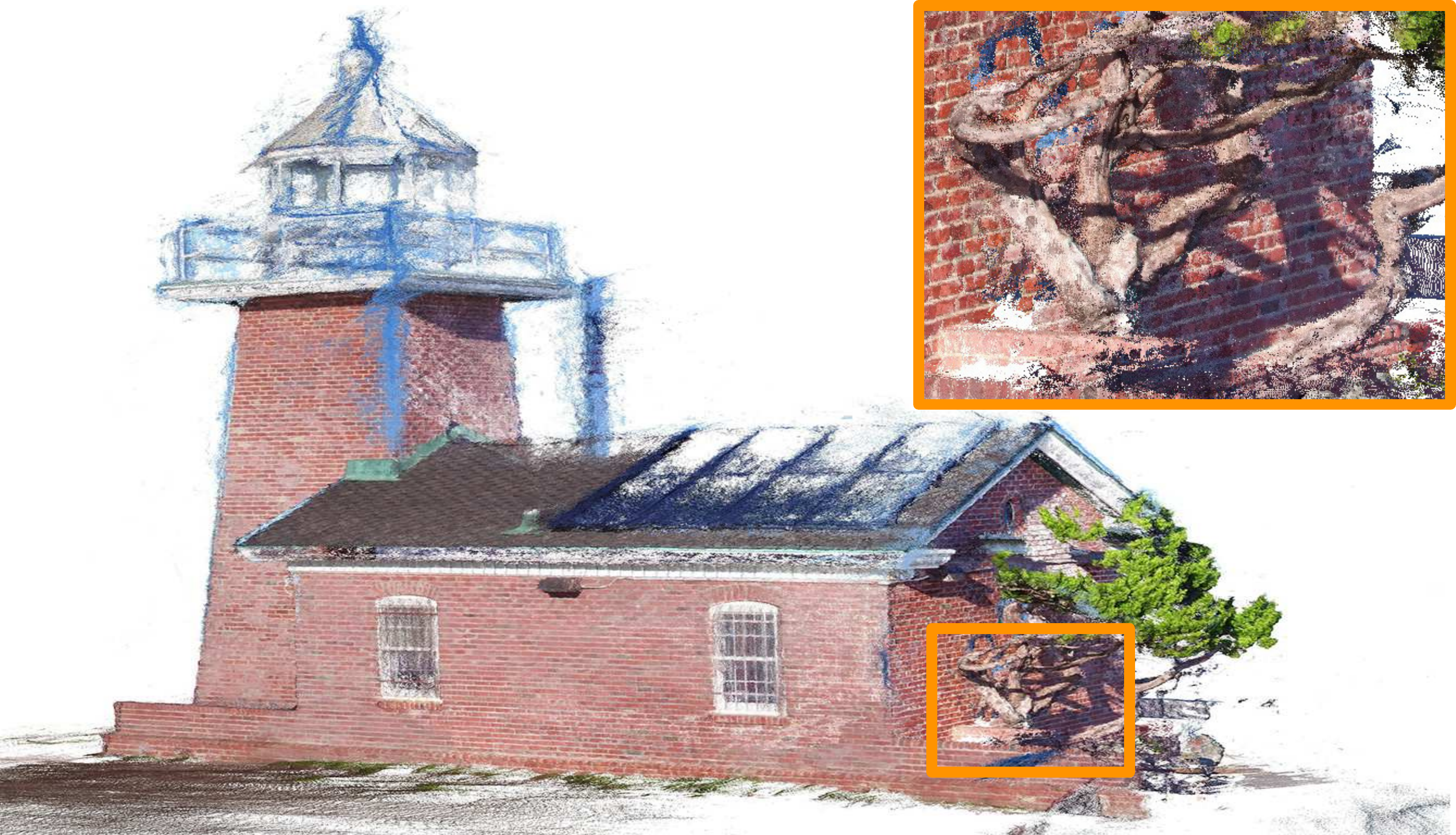}
          & \includegraphics[width=0.226\linewidth]{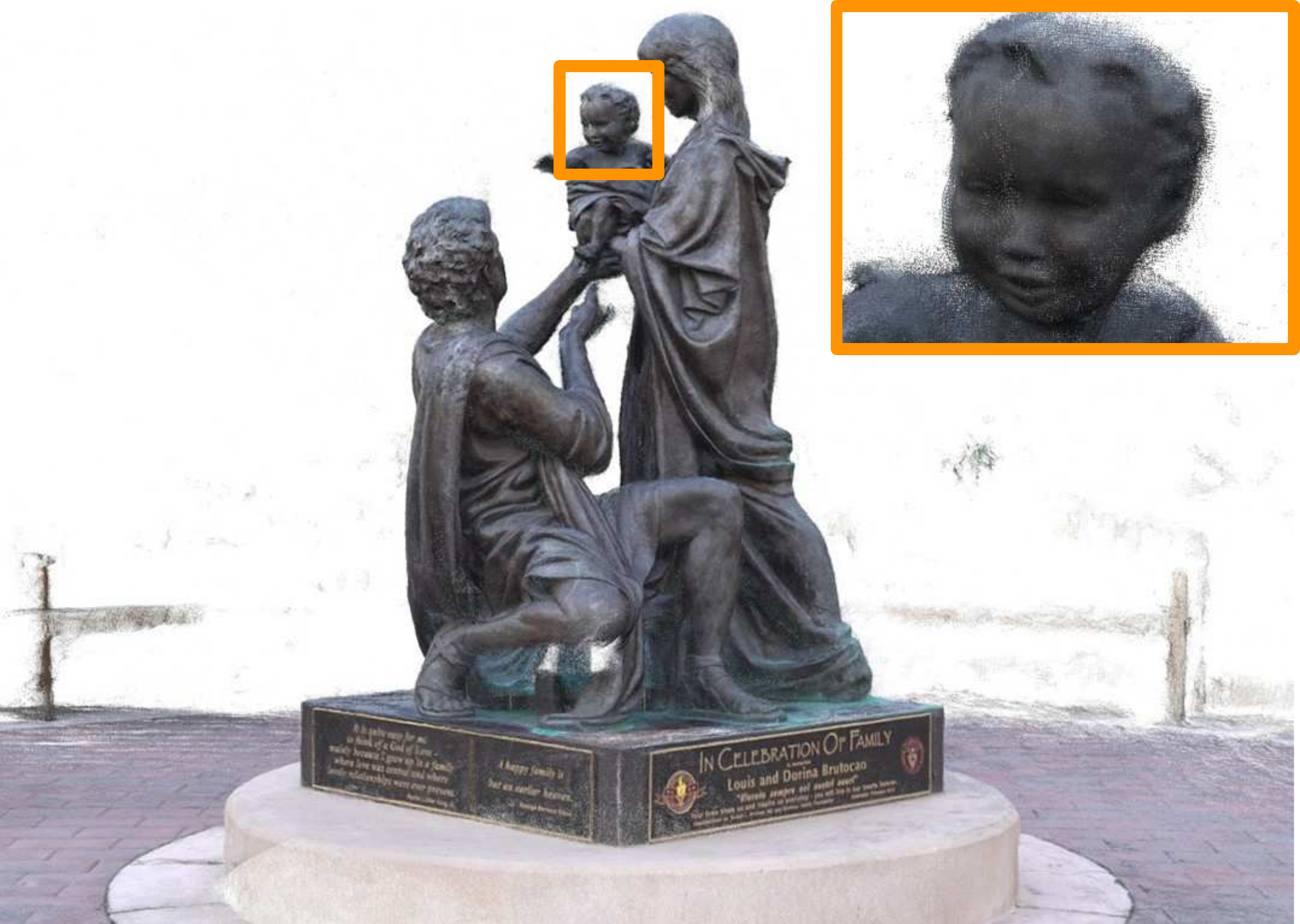}
          & \includegraphics[width=0.284\linewidth]{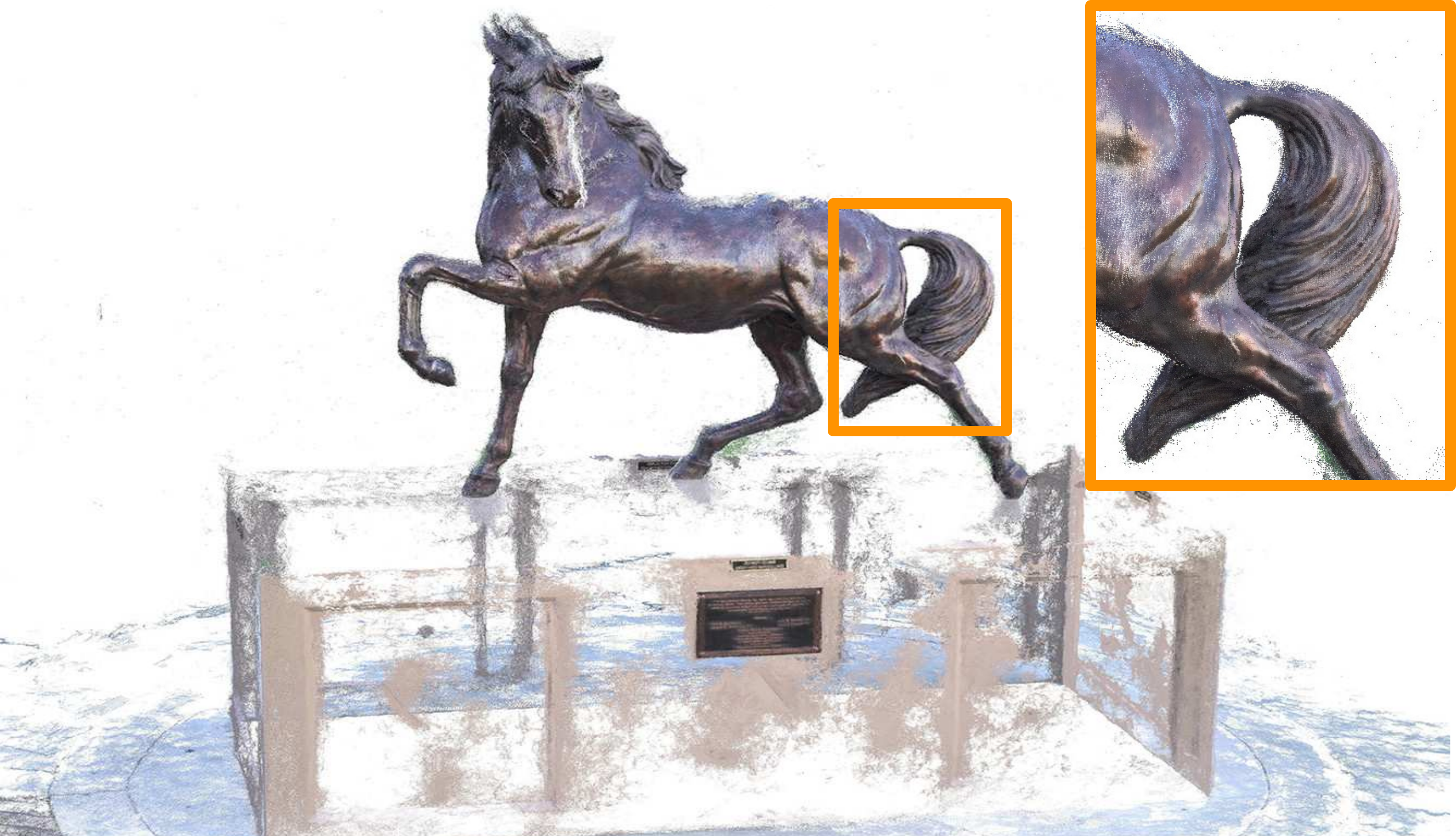}\\
         (c) Lighthouse & (d) Family & (e) Horse
    \end{tabular}
    }
    \end{center}
    \vspace{-0.5cm}
    \caption{Point cloud reconstruction of Tanks and Temples dataset~\cite{knapitsch2017tanks}. Best viewed on screen.}
    \label{fig:qualityTanks}
    \vspace{-0.25cm}
\end{figure*}

\subsection{Results on Tanks and Temples}
\vspace{-0.1cm}
We now evaluate the generalization ability of our method. To this end, we use the model trained on DTU \textbf{without any fine-tuning} to reconstruct point clouds for scenes in Tanks and Temples dataset. For fair comparison, we use the same camera parameters, depth range and view selection of MVSNet~\cite{yao2018mvsnet}. For comparison, we consider four baselines~\cite{chen2019point,luo2019p,yao2018mvsnet,yao2019recurrent} and evaluate the \textit{f-score} on Tanks and Temples. Table~\ref{tanks} summarizes these results. As shown, our method yielded a mean \textit{f-score} 5\% higher than Point-MVSNet~\cite{chen2019point}, which is the best baseline on DTU dataset, and only 1\% lower than P-MVSNet~\cite{luo2019p}. Note that P-MVSNet~\cite{luo2019p} applies more depth filtering process for point cloud fusion than ours which just follows the simple fusion process provided by MVSNet~\cite{yao2018mvsnet}. Qualitative results of our point cloud reconstructions are shown in Fig.~\ref{fig:qualityTanks}. 

\begin{table*}[!t]
\footnotesize
\begin{center}
\begin{tabular}{cc}
\resizebox{0.38\linewidth}{!}{
\begin{tabular}{c|c|ccc}
\hline
Levels & Coarsest Img. Size & Acc. & Comp. & Overall \\
\hline\hline
\comment{1 & 128x160 & 0.483 & 0.526 & 0.5045  \\}
2 & 80x64 & \textbf{0.296} & \textbf{0.406} & \textbf{0.351}  \\
3 & 40x32 & 0.326 & 0.407 & 0.366 \\
4 & 20x16 & 0.339 & 0.411 & 0.375  \\
5 & 10x8  & 0.341 & 0.412 & 0.376 \\
\hline
\end{tabular} 
}
&
\resizebox{0.4\linewidth}{!}{
\begin{tabular}{c|ccc} 
\hline
Pixel Interval & Acc. (mm) & Comp. (mm) & Overall (mm) \\
\hline\hline
2              & 0.299     & 0.413      & 0.356        \\
1              & 0.299     & \textbf{0.403}      & \textbf{0.351}        \\
0.5            & \textbf{0.296}     & 0.406      & \textbf{0.351}        \\
0.25           & 0.313     & 0.482      & 0.397        \\
\hline
\end{tabular} 
} \\(a)&(b)
\end{tabular}
\end{center}
\vspace{-0.55cm}
\caption{Parameter sensitivity on DTU dataset. a) Accuracy as a function of the number of pyramid levels. b) Accuracy as a function of the interval setting.}\label{table:interval}
\vspace{-0.4cm}
\end{table*}

\subsection{Ablation study}
\noindent\textbf{Training pyramid levels.} We first analyze the effect of the number of pyramid levels on the quality of the reconstruction. To this end, we downsample the images to form pyramids with four different levels. Results of this analysis are summarized in Table~\ref{table:interval}a. As shown, the proposed 2-level pyramid is the best. As the level of pyramid increases, the image resolution of the coarsest level decreases. For more than 2-levels, this resolution is too small to produce a good initial depth map to be refined.

\noindent\textbf{Evaluation pixel interval settings.} We now analyze the effect of varying the pixel interval setting for depth refinement. As discussed in section~\ref{sec:depthEstimator}, the depth sampling is determined by the corresponding pixel offset in source views, hence, it is important to set a suitable pixel interval. Table \ref{table:interval}b summarizes the effect of varying the interval from depth ranges corresponding to 0.25 pixel to 2 pixels during \mbox{evaluation}. As shown, the performance drops when the interval is too small (0.25 pixel) or too large (2 pixels).

\section{Conclusion}
In this paper, we proposed CVP-MVSNet, a~\emph{cost volume pyramid} based depth inference framework for MVS. CVP-MVSNet is compact, lightweight, fast in runtime and can handle high resolution images to obtain high quality depth map for 3D reconstruction. Our model achieves better performance than state-of-the-art methods by extensive evaluation on benchmark datasets. In the future, we want to explore the integration of our approach into a learning-based structure-from-motion framework to further reduce the memory requirements for different applications.
~\\
\section*{Acknowledgments}
This research is supported by Australian Research Council grants (DE180100628, DP200102274).

{\small
\bibliographystyle{ieee_fullname}
\bibliography{egbib}
}

\end{document}